\documentclass[runningheads]{llncs}

\usepackage{eccv}

\usepackage{eccvabbrv}

\usepackage{graphicx}
\usepackage{booktabs}

\usepackage[accsupp]{axessibility}  % Improves PDF readability for those with disabilities.

\usepackage{hyperref}

\usepackage{orcidlink}

\usepackage{colortbl}
\usepackage{multirow}
\usepackage{wrapfig}
\usepackage{tabularx}
\usepackage{makecell}
\usepackage{enumitem}
\usepackage{marvosym}
\usepackage{float} % 加载 float 宏包
\definecolor{lightergray}{RGB}{230,230,230}

\begin{document}
	%\setlength{\intextsep}{0pt}
	\title{VCP-CLIP: A visual context prompting model for zero-shot anomaly segmentation} 
	
	\titlerunning{Visual context prompting model}
	
	\author{Zhen Qu\inst{1,2}\orcidlink{0009-0000-2173-612X} \and
		Xian Tao\inst{1,2,3} \textsuperscript{\Letter}\orcidlink{0000-0001-5834-5181} \and
		Mukesh Prasad\inst{4} \orcidlink{0000-0002-7745-9667} \and
		Fei Shen\inst{1,2,3}\orcidlink{0000-0001-9263-4489} \and  \\
		Zhengtao Zhang\inst{1,2,3}\orcidlink{0000-0003-1659-7879} \and
		Xinyi Gong\inst{5}\orcidlink{0000-0002-6515-2836} \and
		Guiguang Ding\inst{6}\orcidlink{0000-0003-0137-9975}}
	
	\authorrunning{Z.~Qu et al.}
	
	\institute{CAS Engineering Laboratory for Intelligent Industrial Vision, Institute of Automation, Chinese Academy of Sciences, Beijing, China \and
		University of Chinese Academy of Sciences, Beijing, China \and 
		CASI Vision Technology CO., LTD., Luoyang, China \and
		University of Technology Sydney, Sydney, Australia \and
		Hangzhou Dianzi University, Hangzhou, China \and  
		Tsinghua University, Beijing, China \\
		\email{\{quzhen2022, taoxian2013, fei.shen, zhengtao.zhang\}@ia.ac.cn} \\ 
		\email{mukesh.prasad@uts.edu.au}, \email{gongxinyi@hdu.edu.cn} \\ 
		\email{dinggg@tsinghua.edu.cn} }
	
	\maketitle

	\begin{abstract}
		Recently, large-scale vision-language models such as CLIP have demonstrated immense potential in zero-shot anomaly segmentation (ZSAS) task, utilizing a unified model to directly detect anomalies on any unseen product with painstakingly crafted text prompts. However, existing methods often assume that the product category to be inspected is known, thus setting product-specific text prompts, which is difficult to achieve in the data privacy scenarios. Moreover, even the same type of product exhibits significant differences due to specific components and variations in the production process, posing significant challenges to the design of text prompts. In this end, we propose a visual context prompting model (VCP-CLIP) for ZSAS task based on CLIP. The insight behind VCP-CLIP is to employ visual context prompting to activate CLIP’s anomalous semantic perception ability. In specific, we first design a Pre-VCP module to embed global visual information into the text prompt, thus eliminating the necessity for product-specific prompts. Then, we propose a novel Post-VCP module, that adjusts the text embeddings utilizing the fine-grained features of the images. In extensive experiments conducted on 10 real-world industrial anomaly segmentation datasets, VCP-CLIP achieved state-of-the-art performance in ZSAS task. The code is available at \url{https://github.com/xiaozhen228/VCP-CLIP}.
	\end{abstract}

	\section{Introduction}
	\label{sec:intro}
	
	In the field of industrial visual inspection, zero-shot anomaly segmentation (ZSAS) endeavors to accurately localize and segment anomalous regions within novel products, without relying on any pre-customized training data. Due to its significant potential applications in scenarios with data privacy concerns or a scarcity of annotated data, ZSAS has garnered increasing attention from researchers \cite{WinCLIP, AnVoL,VAND, AnomalyCLIP}. Unlike traditional anomaly segmentation methods \cite{tao}, ZSAS requires strong generalization ability to adapt to significant variations in visual appearance, anomalous objects, and background features across different industrial inspection tasks.
	\par
\begin{figure}[tb]
		\centering
		\includegraphics[width=0.85\columnwidth]{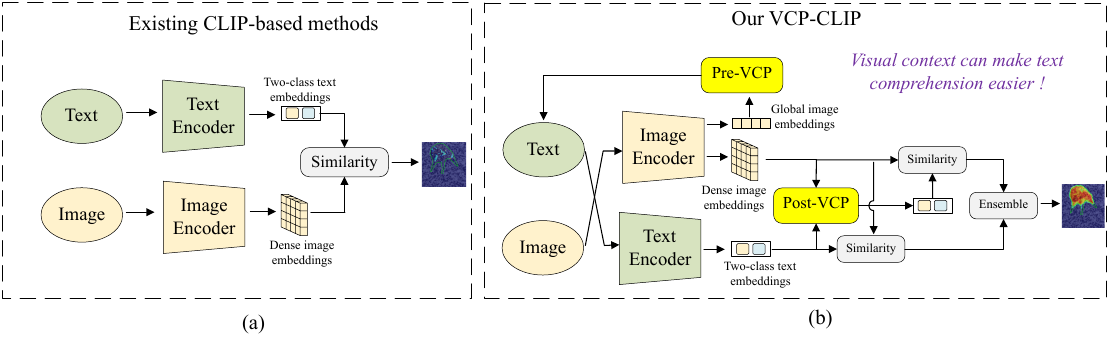}
		\caption{A comparison between existing CLIP-based methods and VCP-CLIP. VCP-CLIP introduces a Pre-VCP module and a Post-VCP module, offering a distinct enhancement over existing CLIP-based methods. (a) Existing CLIP-based methods. (b) VCP-CLIP}
		\label{fig1}
	\end{figure}

 
	In recent, CLIP \cite{CLIP} has emerged as a vision-language foundation model for addressing the ZSAS task. As shown in Fig. \ref{fig1}(a), existing CLIP-based methods map images and their corresponding two-class text into a joint space and compute cosine similarity. Image regions that have high similarity with the defect-related text are considered as anomalies. For example, WinCLIP \cite{WinCLIP}, AnVoL \cite{AnVoL}, and APRIL-GAN \cite{VAND} extract dense visual features by applying multi-scale windowing or patching to images and align normal and abnormal image regions separately through a two-class text prompt design. However, the existing CLIP-based methods \cite{WinCLIP, AnVoL ,VAND, AnomalyCLIP} present significant challenges in practical applications. On the one hand, previous methods \cite{WinCLIP, AnVoL, VAND } assume that the product category (e.g., wood) of inspected images is known in advance and utilize this information to design product-specific textual prompts (e.g., a photo of a normal wood). However, the product categories are unattainable or unpredictable in data privacy scenarios, rendering these methods unusable. Furthermore, we conducted an experiment in which we replaced the product categories (names) in the text prompts with semantically similar terms in WinCLIP, such as substituting \textit{bottle} with \textit{container} or \textit{vessel}. We observed fluctuations in segmentation performance of up to ±8\% in terms of Average Precision (AP) metric. This motivates us to reconsider the importance of product names in text prompts, especially since some product names are ambiguous (e.g., \textit{pcb1}, \textit{pcb2}, \textit{pcb3} in the VisA \cite{VisA} dataset). Even within the same product category, significant differences arise due to specific components and differences in the production process, such as variations in appearance color, size, and manufacturing materials, among others. Recently, AnomalyCLIP \cite{AnomalyCLIP} attempted to design object-agnostic text prompts, but they replaced all product name with a uniform description "object", leading to challenges in adapting to complex industrial scenarios. On the other hand, mapping images and text separately into a joint space \cite{AnomalyCLIP, WinCLIP, VAND} without any interaction does not facilitate mutual understanding of various modalities, and easily leads to image overfitting to certain text prompts. As illustrated in Fig. \ref{fig1}(a), where the output image and text embeddings are directly aligned, this approach results in a limited grasp of diverse modalities, thereby affecting anomaly segmentation performance.
	\par 
\begin{figure}[tb]
		\centering
		\includegraphics[width=0.9\columnwidth]{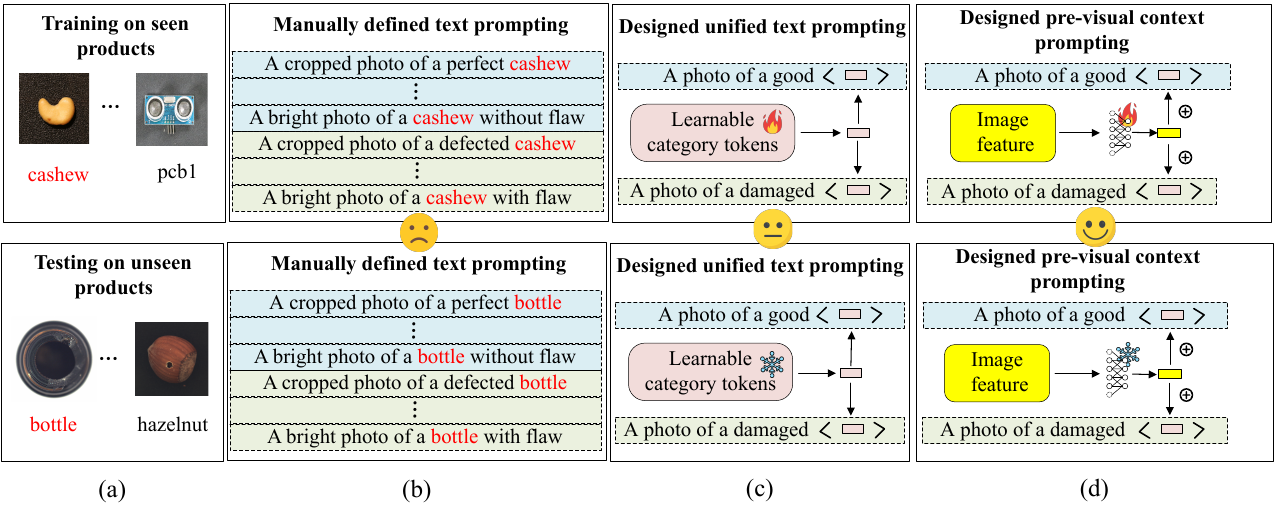}
		\caption{Comparison of different text prompting methods. (a) Task setting. (b) Manually defined text prompting. (c) Designed unified text prompting. (d) Designed pre-visual context prompting.}
		\label{fig2}
	\end{figure}

	To address the aforementioned problems, a straightforward and effective visual context prompting (VCP) model based on CLIP is proposed for ZSAS task. As shown in Fig. \ref{fig2}(a), we aim to perform anomaly segmentation on novel (unseen) products (such as bottle and hazelnut) after training on limited seen products (such as cashews and pcb1) in auxiliary datasets. Existing methods \cite{WinCLIP, VAND} rely on manually defined text prompts as shown in Fig. \ref{fig2}(b). The unified text prompts are used as the baseline as shown in Fig. \ref{fig2}(c) in this paper, where the product categories are set as continuous learnable tokens. The proposed Pre-VCP module, depicted in Fig. \ref{fig2}(d), is an upgraded version of the baseline. It incorporates global image features to more accurately encode the product category semantics in the text space. To facilitate understanding of global image features, a deep text prompting (DTP) technique is introduced to refine the text space. Compared to the baseline, Pre-VCP enables the transition from uniform prompts to image-specific prompts, significantly reducing the cost of prompt designs. To enhance the mutual understanding of features from different modalities, the Post-VCP module is further proposed, which adjusts the output text embeddings based on fine-grained visual features. This approach further strengthens CLIP's ability to accurately segment anomalous regions.
	\par 
	In conclusion, we propose a visual context prompting model based on CLIP (VCP-CLIP) for the ZSAS task. As depicted in Fig. \ref{fig1}(b), we extract the global and dense image embeddings from the image encoder. The former is integrated into the input text prompts after passing through the Pre-VCP module, while the latter is utilized for fine-grained image features in anomaly segmentation. A Post-VCP module is further designed to update the text embeddings based on fine-grained visual features, effectively facilitating mutual understanding between different modalities and further enhancing the model's generalization ability to novel products. The final anomaly maps simultaneously integrate segmentation results aligned from the original text embeddings and dense image embeddings, which helps further enhance the segmentation performance.
	\par 
	The main contributions of this work are as follows:
	\par
	1. We propose a novel visual context prompting model based on CLIP, namely VCP-CLIP, to tackle the ZSAS problem. By training on a limited set of seen products, VCP-CLIP can localize anomalies in any unseen product, even when the product category is unknown. Compared to current text prompting approaches \cite{WinCLIP, AnVoL, AnomalyCLIP, VAND}, our approach utilizes visual context prompting to fully activate CLIP's anomalous semantic perception ability.
	\par 
	2. We reveal for the first time that visual context provides additional information for text prompts in the ZSAS task. Specifically, the Pre-VCP and Post-VCP modules are designed to utilize global and fine-grained image features for text prompting, respectively. In doing so, VCP-CLIP avoids extensive manually defined text prompting engineering, thus alleviating the overfitting issue arising from pre-training on specific text prompts. 
	\par 
	3.	In extensive experiments conducted on 10 real-world industrial anomaly segmentation datasets, VCP-CLIP exhibits superior zero-shot performance in segmenting anomalies on unseen products.

	\section{Related Work}
	
	\textbf{Prompt learning.} Prompt learning is initially applied in the field of NLP, aiming to utilize affordable annotated data to automatically generate prompts, thereby enhancing the capabilities of foundation models, such as CLIP \cite{CLIP}, GPT-3.5 \cite{GPT}, and LLaMA \cite{llama} in downstream tasks. CoOp \cite{CoOp} first introduces prompt learning in the CLIP model, utilizing learnable prompt tokens in the textual space. VPT \cite{VPT} and ZegCLIP \cite{ZigCLIP} insert trainable embeddings in each layer of the image encoder, allowing refinement of the image space to better adapt to downstream semantic segmentation task. These methods aim to enable the pretrained backbone to adapt to the target domain using prompt learning. In recent works, CoCoOp \cite{CoCoOp} and DenseCLIP \cite{DenseCLIP} guide the pretrained backbone to adapt to the target domain through the visual context prompting. Related to our VCP module is CoCoOp, which incorporates visual contexts into text prompts to improve the classification performance on novel categories. However, our VCP replaces product categories within the text prompts rather than the entire sentence, in contrast to CoCoOp. The proposed approach has been validated as more effective than CoCoOp in ZSAS, which does not necessitate prior knowledge of product categories.  
	\par 
	\textbf{Zero-shot anomaly segmentation.} With the advancements of foundation models such as CLIP \cite{CLIP} and SAM \cite{SAM}, ZSAS has increasingly captured the attention of researchers. According to whether auxiliary data for training is required, existing methods can be broadly categorized into two groups. 1) Training-free methods. Building upon CLIP, WinCLIP \cite{WinCLIP} and AnVoL \cite{AnVoL} carefully craft text prompts to identify anomalies without training on auxiliary datasets. The former proposes a window-based approach, aggregating classification results from images within different scale windows using harmonic aggregation. The latter utilizes V-V attention instead of the original Q-K-V attention in the image encoder to extract fine-grained features and adaptively adjusts for each image during testing in a self-supervised manner. SAA/SAA+ \cite{SAA} utilizes language to guide the Grounding DINO \cite{grounding} for detection of anomalous regions and then employs SAM for finely segmenting the detection results. However, these existing methods not only require more complex prompt designs or post-processing but also introduce additional computational and storage burdens during inference. 2) Training-required methods. APRIL-GAN \cite{VAND}, CLIP-AD \cite{CLIPAD}, and AnomalyCLIP \cite{AnomalyCLIP} utilize seen products with annotations as auxiliary data to fine-tune CLIP for ZSAS on unseen products. These approaches employ linear layers to map patch-level image features to a joint space of text and vision, facilitating alignment between different modalities. AnomalyGPT \cite{AnomalyGPT} is another seminal work that utilizes the large language model Vicuna \cite{vicuna} to guide the model in locating anomalies. Through supervised pretraining on synthesized anomaly images, AnomalyGPT can support multi-turn dialogues and locate anomalies in unseen products. However, existing methods all overlook the role of visual context in fine-grained multimodal alignment, and they may struggle when confronted with complex industrial anomaly segmentation scenes. Recently, ClipSAM \cite{CLIPSAM}, an integration of CLIP and SAM, has been employed for cross-modal interaction in ZSAS task. However, the two-stage prediction has increased the complexity of the model.
	\section{Our Method}
	\begin{figure}[tb]
		\centering
		\includegraphics[width=0.95\columnwidth]{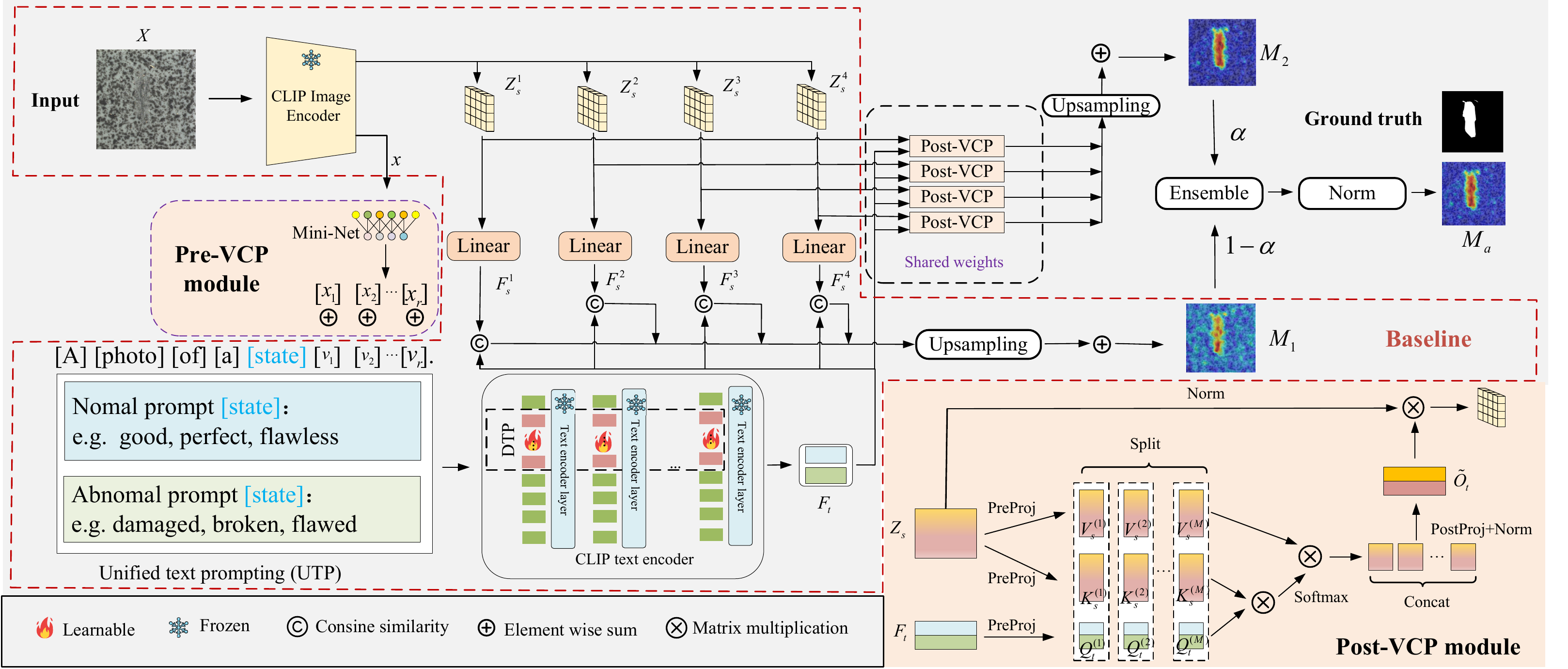}
		\caption{Framework of VCP-CLIP. Our approach incorporates richer visual knowledge into the textual space, and cross-modal interaction between textual and visual features by using a Pre-VCP module and a Post-VCP module.}
		\label{fig3}
	\end{figure}
	\subsection{Problem Definition}
	Our approach follows the generalized ZSAS methods adopted in works \cite{VAND, AnomalyCLIP}, which requires segmenting the anomalies in unseen products $C^{u}$ after training on seen products $C^{s}$ with pixel-annotations. During the training stage, the model generates pixel-wise classification results based on two categories of textual descriptions: normal and abnormal. During the testing stage, the model is expected to directly segment anomalies in unseen products. It is worth noting that $C^u \cap C^s = \emptyset$ and the products used in the training and testing stages come from different datasets. This undoubtedly poses a significant challenge to the model's domain generalization capability.
	
	\subsection{The design of baseline}
	Existing CLIP-based improvement methods have three main drawbacks: 1) manually designing text prompts is time-consuming and labor-intensive, 2) product-specific text prompts cannot adapt to data privacy scenarios, and 3) the localization results are easily influenced by the semantics of product categories in the text prompts \cite{AnomalyCLIP}. To address the aforementioned issues, we propose a baseline that incorporates two main designs: unified text prompting (UTP) and deep text prompting (DPT). As shown in Fig. \ref{fig3}, given an input image $X \in \mathbb{R}^{h\times w \times 3}$ and two-class text prompts, the designed baseline (marked in red dashed) first extracts patch-level image features and text features separately. Then, the patch-level image features are mapped to a joint space, where the similarity between image features and text features is computed to generate anomaly maps. Finally, anomaly maps from multiple intermediate layers of the image encoder are fused after upsampling to obtain the final results.
	\par
	\textbf{Unified text prompting (UTP).} A unified template for generating normal and abnormal text prompts is designed as follows:
	\begin{equation*}
		\label{eq16}
		H = [a] [photo] [of] [a] [state] [v_1] [v_2] \cdots [v_r]
	\end{equation*}
	where $v_i, i\in\{1,2,\cdots r\}$ is a $C$-dimensional learnable vector embedded into the word embedding space, used to learn the unified textual context of the product categories. A pair of opposing [state] words, such as "good/damaged" and "perfect/flawed", is utilized to generate normal and abnormal text prompts, respectively. $H$  represents the word embedding matrix corresponding to specific prompts in the textual space. In this paper, we choose a common state word pair, i.e. "good/damaged".
	\par
	\textbf{Deep text prompting (DTP).} Before statement, let us first review the inference process of the CLIP text encoder briefly. Before being fed into the text encoder, [SOS] and [EOS] are respectively added to the front and back of the text prompt, indicating the beginning and end of the sentence. Afterwards, these tokens are mapped to a discrete word embedding space, capped to a fixed length of 77 in CLIP. Let us denote the word embeddings as $[s, H, e, J] \in \mathbb{R}^{77\times C}$, where $s$ and $e$ are $C$-dimensional word embeddings corresponding to [SOS] and [EOS] tokens, respectively. $J$ is a placeholder matrix initialized to zero to ensure a fixed length of the word embeddings. The final output text embedding at the position of the [EOS] token is aligned with the image features after passing through a linear projection layer. 
	\par  
	To better align fine-grained normal and anomalous visual semantics with text, deep text prompting is designed to further refine the textual space as shown in Fig. \ref{fig3}. In specific, continuous trainable embeddings are inserted at the beginning of text embedding in each transformer layer of the text encoder. Assuming the text encoder's (i+1)-th layer is represented as $Layer_{i+1}^{text}$, the inserted embeddings are $P_i \in \mathbb{R}^{n \times C}$ and the output text embedding is $g$. The process is formulated as follows:
	\begin{gather}
		[s_i, \_ , H_i, e_i, J_i] = Layer_i^{text}([s_{i-1}, P_{i-1}, H_{i-1}, e_{i-1}, J_{i-1}])  \\
		g = TextProj (Norm(e_{N_t}))
	\end{gather}
	where $i = 1,2,\cdots N_t$, $s_0 = s$, $H_0 = H$, $e_0 = e$. $N_t$ is the number of text encoder layers. $TextProj(\cdot)$ and $Norm(\cdot)$ respectively denote final text projection and LayerNorm \cite{LayerNorm} layers. For normal and abnormal text prompts, we denote the embeddings after DTP as $g_n$ and $g_a$, respectively. Since the masked self-attention is employed in the text encoder, $[s_i, P_i, H_i, e_i, J_i]$ and $[s_i, H_i,P_i, e_i, J_i]$ are not mathematically equivalent. We adopted the former because the model can only attend to tokens before itself, thus placing the learnable embeddings at the beginning of the sentence leads to a greater degree of refinement in the textual space. More details are shown in the Appendix B.2.
	
	\textbf{How to acquire the anomaly map?} For an input image $X \in \mathbb{R}^{h\times w \times 3}$, patch-level visual feature map $Z_s^l \in \mathbb{R}^{H\times W \times d_I}, l=1,2,\cdots, B$ are extracted from the image encoder layers, where $H = h/patchsize, W=w/patchsize$, $d_I$ is the size of image embeddings and $B$ is the number of extracted intermediate patch-level feature layers. Then, the feature maps are mapped to a joint space and align with text embeddings using a single linear layer by calculating the cosine similarity. Let us respectively denote the visual and textual features in the joint space as $F_s^l \in \mathbb{R}^{HW \times C}$ and $F_t = [g_n,g_a] \in \mathbb{R}^{2\times C}$, where $C$ is the embedding size in the joint space. The process of acquiring the anomaly map can be formulated as:
	\begin{equation}
		\label{eq16}
		M_1^l = softmax(Up(\widetilde{F}_s^l\widetilde{F}_t^T) / \tau_1), l=1,2,\cdots B
	\end{equation}
	where $\tau_1$ denotes the temperature coefficient, which is set as a learnable parameter. $Up(\cdot)$ is an upsampling operation with bilinear interpolation. $\widetilde{(\cdot)}$ represents the $L_2$-normalized version along the embedding dimension.
	\par
	\subsection{The design of VCP-CLIP}
	The baseline has made some progress, but still faces the following three main problems: 1) The unified text prompt does not consider specific visual contexts. 2) Overfitting phenomena may occur in the unified text prompt. 3) Insufficient interaction between information from different modalities limits further improvement in segmentation performance. In this end, we further designed two novel visual context prompting modules, namely Pre-VCP and Post-VCP as shown in Fig. \ref{fig3}. In contrast to the baseline, the global features of the image are encoded into the text prompt using the Pre-VCP module. The Post-VCP module receives patch-level features from the image encoder and text features from the text encoder as inputs to generate the anomaly map.
	\par 
	\textbf{Pre-VCP module.} We designed a Pre-VCP module to introduce global image features into the text prompts of the baseline. Due to the extensive alignment of image-text pairs during the pretraining process of CLIP, the embedding at the [CLS] token position of the image encoder encompasses rich global image features. We combine the global image features with learnable vectors in the baseline to facilitate the fusion with the unified category contexts. Specifically, the global image features are initially mapped to the word embedding space through a small neural network, namely \textit{Mini-Net}. This can be expressed as $\{x_i\}_{i=1}^{r} = h(x)$, where $x_i \in \mathbb{R}^{1\times C}, i=1,2,\cdots r$ represents the mapping results, which are combined with embeddings corresponding to the product category:
	\begin{equation}
		\label{eq16}
		z(x,v) = [z_1(x_1,v_1), z_2(x_2,v_2), \cdots , z_r(x_r,v_r)]
	\end{equation}
	where $z_i = x_i + v_i$. For the \textit{Mini-Net} $h(\cdot)$, a parameter-efficient design utilizing only a one-dimensional convolutional layer with ($r$, $1\times 3$) kernels is employed.
	The final text prompt based on Pre-VCP can be expressed as follows:
	\begin{equation*}
		\label{eq16}
		H_v = [a][photo][of][a][state][[z_1(x_1,v_1)][z_2(x_2,v_2)]\cdots [z_r(x_r,v_r)]
	\end{equation*}
	For convenience in the subsequent text, we refer to the text prompt template as "a photo of a [state] [$z(x,v)$]".
	\par
	\textbf{Post-VCP module.} To further enable the text embedding to adapt based on fine-grained image features, we devised a Post-VCP module, as illustrated in Fig. \ref{fig3}. The text embedding $F_t \in \mathbb{R}^{2\times C}$  and flattened visual embedding $Z_s^l \in \mathbb{R}^{HW \times d_I}$ from each layer are projected into a latent space with $C$-dimension. Then the learnable queries $Q_t$, keys $K_s^l$, and values $V_s^l$ can be obtained:
	\begin{equation}
		\label{eq16}
		Q_t = F_tW_t^q, K_s^l = Z_s^l W_s^k, V_s^l = Z_s^lW_s^v
	\end{equation}
	where $W_t^q\in\mathbb{R}^{C\times C},W_s^k\in\mathbb{R}^{d_I\times C},W_s^v\in\mathbb{R}^{d_I\times C}$ are linear projection matrices in the \textit{PreProj} layer. To capture richer visual features for fine-tuning text, a multi-head structure is adopted for computing attention maps to update text features within each head using matrix multiplication: 
	\begin{gather}
		\{Q_t^{(m)}\}\{K_s^{l(m)}\}\{V_s^{l(m)}\} = Split(Q_t, K_s^l, V_s^l)  \\
		A_t^{l(m)} = SoftMax(Q_t^{(m)}K_s^{l(m)T}),\quad O_t^{l(m)} = A_t^{l(m)}V_s^{l(m)} \\
		O_t^l = Concat(O_t^{l(1)},O_t^{l(2)},\cdots,O_t^{l(M)})W_t^o
	\end{gather}
	where $m = 1,2,\cdots, M$. $M$ is the number of heads, $Q_t^{(m)} \in \mathbb{R}^{2\times(C/M)}, K_s^{l(m)}  \in \mathbb{R}^{HW\times(C/M)},V_s^{l(m)} \in \mathbb{R}^{HW\times(C/M)}$ represent the features within each head after the $Split(\cdot)$ operation for partitioning along the embedding dimension. $A_t^{l(m)} \in \mathbb{R}^{2\times HW}$ and $O_t^{l(m)}\in \mathbb{R}^{2\times (C/M)}$ respectively refer to the attention maps and the text features updated through the image feature within each head. After concatenating all features along the embedding dimension using the $Concat(\cdot)$ operation, a \textit{PostProj} layer with weight matrix $W_t^o \in \mathbb{R}^{C\times d_I}$ is employed to obtain the final updated text embedding $O_t^l \in \mathbb{R}^{2\times d_I}$ from $F_t$. Then, the updated anomaly map is calculated as:
	\begin{equation}
		\label{eq16}
		M_2^l = softmax(Up(\widetilde{Z}_s^l{\widetilde{O}_t}^{lT}) / \tau_2), l=1,2,\cdots B
	\end{equation}
	where $\tau_2$ is a temperature coefficient set as a learnable parameter. 
	\par
	\begin{wrapfigure}{r}{0.5\textwidth} 
		\centering
		%\setlength{\intextsep}{0pt}
		%\vspace{-25pt}
		%\hspace{-25pt}
		\includegraphics[width=0.4\textwidth]{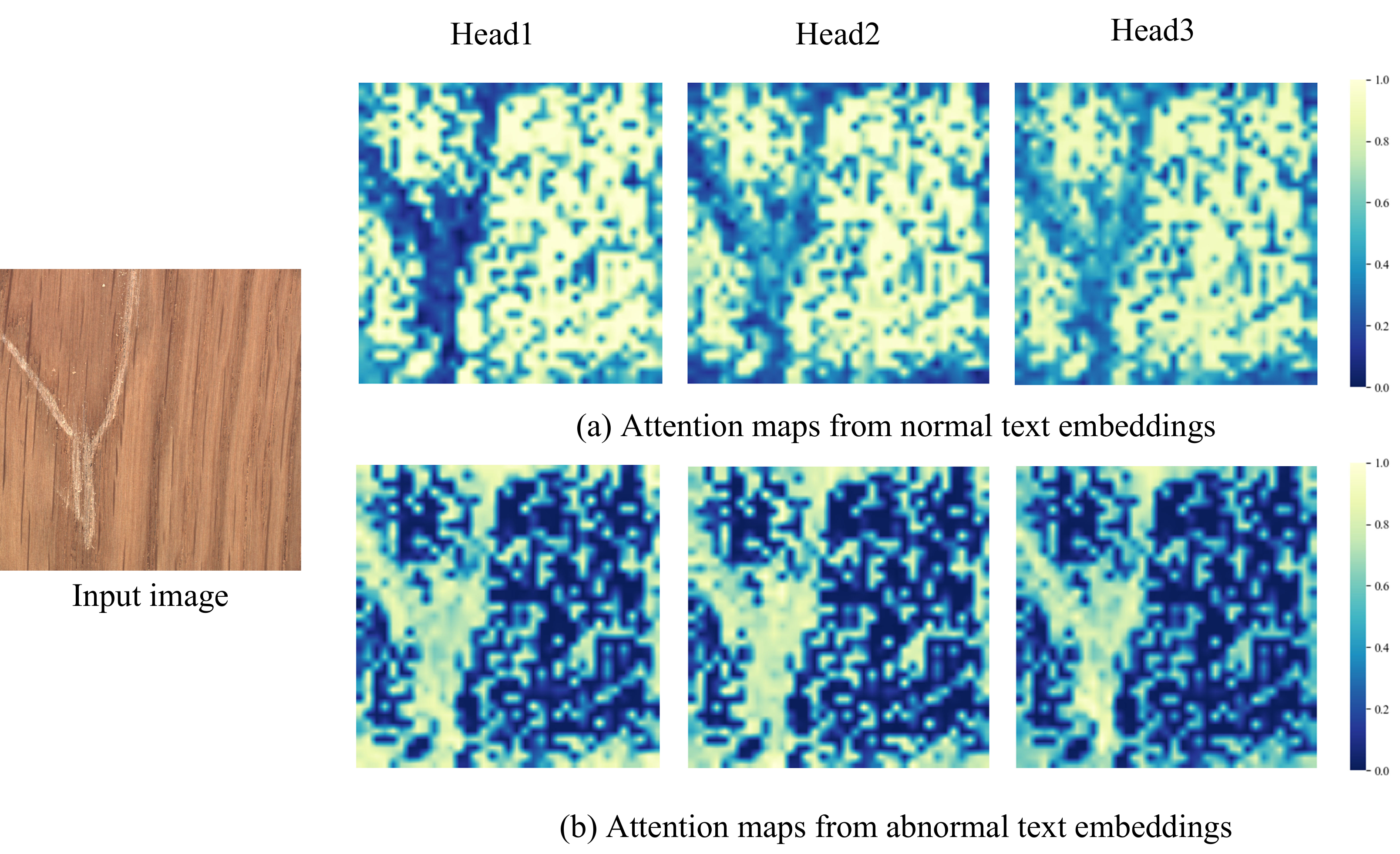} 
		\caption{The visualization result of the attention maps from the Post-VCP module.}
		\label{fig4}
	\end{wrapfigure}
	To visually validate the effectiveness of the Post-VCP module, we show the attention maps $A_t^{l(m)}$ under different heads corresponding to normal and abnormal text embeddings. These maps reveal that abnormal text embeddings concentrate more on defective regions of the image compared to normal text embeddings. This clear differentiation stems from employing fine-grained visual contexts in the Post-VCP module to update text embeddings from $F_t$ to $O_t^l$.
	
	\subsection{Training and Inference}
	\textbf{Loss function.} In this work, we employed focal loss \cite{focal} and dice loss \cite{dice} to supervise the learning of VCP-CLIP. The total loss function of VCP-CLIP is calculated as:
	\begin{equation}
		\begin{split}
			\label{eq16}
			L_{\text{total}} &= \underbrace{\sum_{l} \text{Focal}(M_1^l,S) + \sum_{l} \text{Dice}(M_1^l,S)}_{\text{Baseline}} \\
			&\quad + \underbrace{\sum_{l} \text{Focal}(M_2^l,S) + \sum_{l} \text{Dice}(M_2^l,S)}_{\text{Additional VCP modules}}
		\end{split}
	\end{equation}
	where the loss function consists of two components, one for the baseline and the other for additional VCP module. $M_1^l$ and $M_2^l, l=1,2,\cdots B$ are anomaly maps generated from the two branches mentioned above. $S\in \mathbb{R}^{h\times w}$ is the ground truth corresponding to the input image.
	\par
	\textbf{Inference.} The ultimate anomaly maps come from different layers of the image encoder by summation. The anomaly maps generated from the two branches are represented as $M_1$ and $M_2$. To further enhance the ZSAS capability, we introduced a weighted fusion policy to generate the final anomaly map, $M_a = (1 - \alpha)M_1 + \alpha M_2$ , where $\alpha \in [0,1]$ is a fusion weight designed as a hyperparameter to balance the importance of different anomaly maps.
	\section{Experiments}
	
	\subsection{Experimental Setup}
	\textbf{Datasets and metrics.} To assess the performance of the model, ten real industrial anomaly segmentation datasets are used, including MVTec-AD \cite{MVTec}, VisA \cite{VisA}, BSD \cite{BSD}, GC \cite{GC}, KSDD2 \cite{KSDD2}, MSD \cite{MSD}, Road \cite{Road}, RSDD \cite{RSDD}, BTech \cite{BTech}, DAGM \cite{DAGM}. Since the products in VisA do not overlap with those in other datasets, we use VisA as the training dataset for evaluation on other datasets. For VisA itself, we assess it after training on MVTec-AD. Please refer to the Appendix C for more details. To ensure a fair comparison, pixel-level AUROC (Area Under the Receiver Operating Characteristic), PRO (Per-Region Overlap), and AP (Average Precision) are employed as the evaluation metrics, following the recent works \cite{VAND, CLIPAD}.
	\par 
	\begin{table}[t]
		\caption{Comparison with existing state-of-the-art methods. The (a, b, c) represents the pixel-level AUROC (\%), PRO (\%) and AP (\%), respectively. The methods denoted by $\dag$ are training-free, while the others are training-required.}
		\centering
		\label{tab1}
		\renewcommand{\arraystretch}{1.2}
		\resizebox{1\columnwidth}{!}
		{
			\begin{tabular}{>{\centering\arraybackslash}p{1.8cm} *{5}{>{\centering\arraybackslash}p{2.6cm}}>{\columncolor{lightergray}\centering\arraybackslash}p{2.6cm}>{\columncolor{lightergray}\centering\arraybackslash}p{2.7cm}}
				\toprule
				Datasets & WinCLIP $\dag$ \cite{WinCLIP}            & AnVoL $\dag$ \cite{AnVoL}             & CoCoOp \cite{CoCoOp}             & AnomalyGPT \cite{AnomalyGPT}         & APRIL-GAN  \cite{VAND}        & Baseline(ours)     & VCP-CLIP(ours)     \\   \midrule
				MVTec-AD & (85.1, 64.6, 18.2) & (\underline{90.6}, 77.8, 28.1) & (88.2, 83.2, 40.4) & (79.5, 45.9, 23.7) & (87.6, 44.0, 40.8) & (89.2, \underline{85.8}, \underline{45.2}) & (\textbf{92.0}, \textbf{87.3}, \textbf{49.4}) \\
				VisA     & (79.6, 56.8, 5.4)  & (91.4, 75.0, 12.7) & (94.9, 88.0, 24.8)  & (90.3, 61.5, 13.3) & (94.2, 86.8, 25.7) & (\underline{95.5}, \underline{89.6}, \underline{27.3}) & (\textbf{95.7}, \textbf{90.7}, \textbf{30.1}) \\
				BSD      & (87.7, 56.8, 4.4)  & (96.3, 72.6, 13.3) & (98.7, 85.3, 55.5)  & (87.8, 54.0, 37.9) & (98.8, 61.6, \underline{59.7}) & (\underline{99.1}, \underline{86.4}, 58.5) & (\textbf{99.3}, \textbf{87.0}, \textbf{70.2}) \\
				GC       & (71.9, 44.2, 8.6)  & (92.1, 66.5, 14.1) & (96.1, \underline{81.6}, \underline{41.5})  & (60.0, 11.6, 2.3)  & (94.0, 21.5, 34.4) & (\underline{97.5}, 81.2, 39.6) & (\textbf{97.8}, \textbf{83.8}, \textbf{42.6}) \\
				KSDD2    & (89.4, 65.9, 17.5) & (95.9, 80.4, 33.9) & (96.1, 90.9, 69.6)  & (91.5, 61.9, 29.7) & (97.5, 49.6, 67.2) & (\underline{99.4}, \underline{95.4}, \underline{71.6}) & (\textbf{99.5}, \textbf{98.0}, \textbf{75.2}) \\
				MSD      & (47.0, 41.7, 1.5)  & (95.0, 68.6, 9.4)  & (96.1, 82.3, 27.0)  & (67.9, 22.7, 1.8)  & (98.1, 36.8, 36.0) & (\underline{98.5}, \underline{91.0}, \underline{54.9}) & (\textbf{99.0}, \textbf{91.1}, \textbf{61.0}) \\
				Road     & (78.1, 37.9, 11.0) & (85.8, 39.9, 18.3) & (91.0, 56.0, 29.4)  & (67.6, 15.5, 9.2)  & (89.0, 6.1, \underline{30.4})    & (\underline{92.7}, \underline{62.9}, 30.2) & (\textbf{93.6}, \textbf{66.4}, \textbf{32.1}) \\
				RSDD     & (91.4, 63.6, 3.7)  & (94.7, 75.5, 3.5)  & (99.1, 94.4, 37.4)  & (93.2, 58.4, 16.0) & (99.1, 62.9, \underline{35.9}) & (\underline{99.3}, \underline{95.9}, 35.0) & (\textbf{99.5}, \textbf{97.5}, \textbf{44.1}) \\
				BTech    & (63.2, 22.8, 11.4) & (85.6, 45.4, 32.1) & (90.8, 70.1, 44.4)  & (75.9, 29.3, 17.6) & (90.8, 18.8, 43.6) & (\underline{91.2}, \underline{68.9}, \underline{43.7}) & (\textbf{94.1}, \textbf{74.6}, \textbf{51.4}) \\
				DAGM     & (75.1, 43.1, 3.2)  & (83.4, 64.7, 10.7) & (98.0, 94.7, 42.9)  & (81.9, 35.7, 4.7)  & (99.0, 44.1, \underline{50.5}) & (\underline{99.1}, \underline{97.2}, 48.9) & (\textbf{99.4}, \textbf{98.3}, \textbf{52.0})
				\\   \bottomrule
			\end{tabular}
		}
	\end{table}
	\textbf{Implementation details.} In the experiments, we adopt the CLIP model with ViT-L-14-336 pretrained by OpenAI \cite{CLIP} by default. Specifically, we set the number of layers $B$ for extracting patch-level features to 4. Since the image encoder comprises 24 transformer layers, we evenly extract image features from layers \{6, 12, 18, 24\}. All images are resized to a resolution of $518\times518$, and then fed into the image encoder. The length of the learnable category vectors $r$ and the length of the learnable text embeddings $n$ in each text encoder layer are set to 2 and 1, respectively, by default. The number of attention heads $M$ in the Post-VCP module is set to 8. The fusion weight $\alpha$ for different anomaly maps is set to 0.75 as the default value. The Adam optimizer \cite{Adam} with an initial learning rate of 4e-5 is used, and the model is trained for continuous 10 epochs with a batch size of 32. All experiments are conducted on a single NVIDIA GeForce RTX 3090. We conducted three runs using different random seeds and then averaged the results. More details can be found in Appendix A.
	
	\subsection{Comparison with the State-of-the-art}
	Two kinds of state-of-the-art approaches are used to compare with ours: training-free approaches and training-required approaches. The training-free approaches include WinCLIP \cite{WinCLIP} and AnVoL \cite{AnVoL}, which do not require auxiliary datasets for fine-tuning the model but necessitate more complex manual prompt designs and inference processes. The training-required approaches comprise CoCoOp \cite{CoCoOp}, AnomalyGPT \cite{AnomalyGPT} and APRIL-GAN \cite{VAND}, which adhere to the protocol of training on the seen products and testing on the unseen products. 
	\par 
	\begin{figure}[t]
		\centering
		\includegraphics[width=0.9\columnwidth]{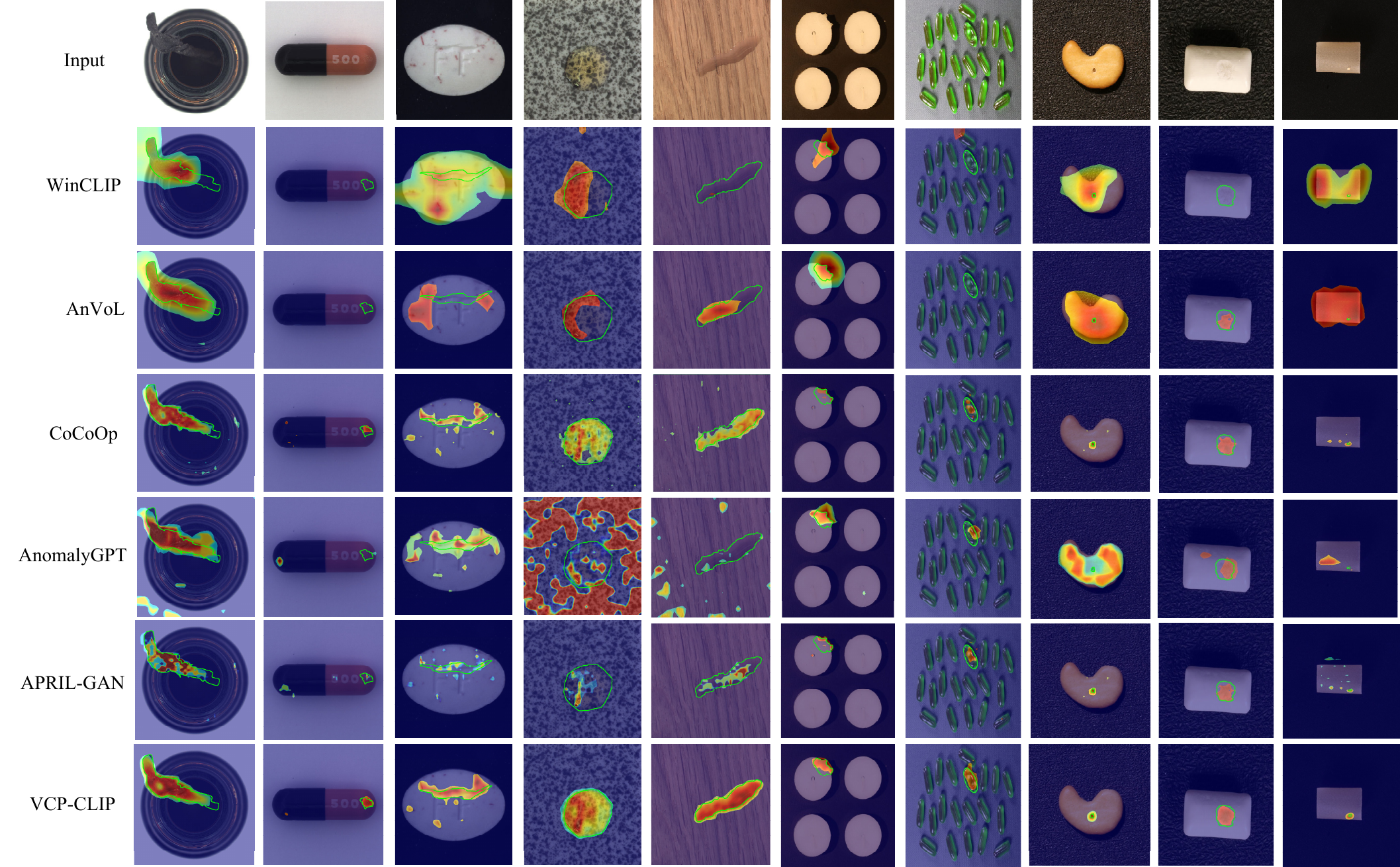}
		\caption{Qualitative segmentation results. The first five columns use images from the MVTec-AD dataset, and the last five are from the VisA dataset.}
		\label{fig5}
	\end{figure}
	\textbf{Quantitative comparison.} Table \ref{tab1} shows the quantitative performance comparison with other state-of-the-art methods on ZSAS. The best results are shown in bold, and the second best results are underlined. It can be observed that the proposed VCP-CLIP outperforms all other methods across all metrics, particularly in terms of AP. Due to the tiny anomaly regions on the Visa dataset, its anomaly segmentation is more challenging. However,  VCP-CLIP still maintains its advantage compared to other methods. Notably, it achieves state-of-the-art results on VisA dataset, with AUROC score of 95.7\%, PRO score of 90.7\% and AP score of 30.1\%. It is noteworthy that our baseline approach has already achieved nearly superior performance compared to existing methods such as CoCoOp, which similarly introduces global image information in the text prompts. This is because our method simultaneously adjusts text embeddings using fine-grained image features.
	\par 
	\textbf{Qualitative comparison.} For a more intuitive understanding of the results, we visualized the anomaly segmentation results of our VCP-CLIP alongside another five methods: WinCLIP \cite{WinCLIP}, AnVoL \cite{AnVoL}, CoCoOp \cite{CoCoOp}, AnomalyGPT \cite{AnomalyGPT}, and APRIL-GAN \cite{VAND} on the MVTec-AD and VisA datasets in Fig. \ref{fig5}. The visualization results clearly indicate that the compared approaches have a tendency to generate incomplete or false-positive results, which can negatively impact the performance of anomaly localization. In contrast, our VCP-CLIP effectively mitigates these issues, providing a more accurate and reliable approach to ZSAS. More quantitative and qualitative comparisons are provided in the Appendix D.
	\par 
	\subsection{Unified text prompting vs. visual context prompting}
\begin{figure}[t]
		\begin{minipage}[b]{0.64\textwidth}
			\centering
			\begin{subfigure}[b]{0.5\textwidth}
				\centering
				\includegraphics[width=0.98\textwidth]{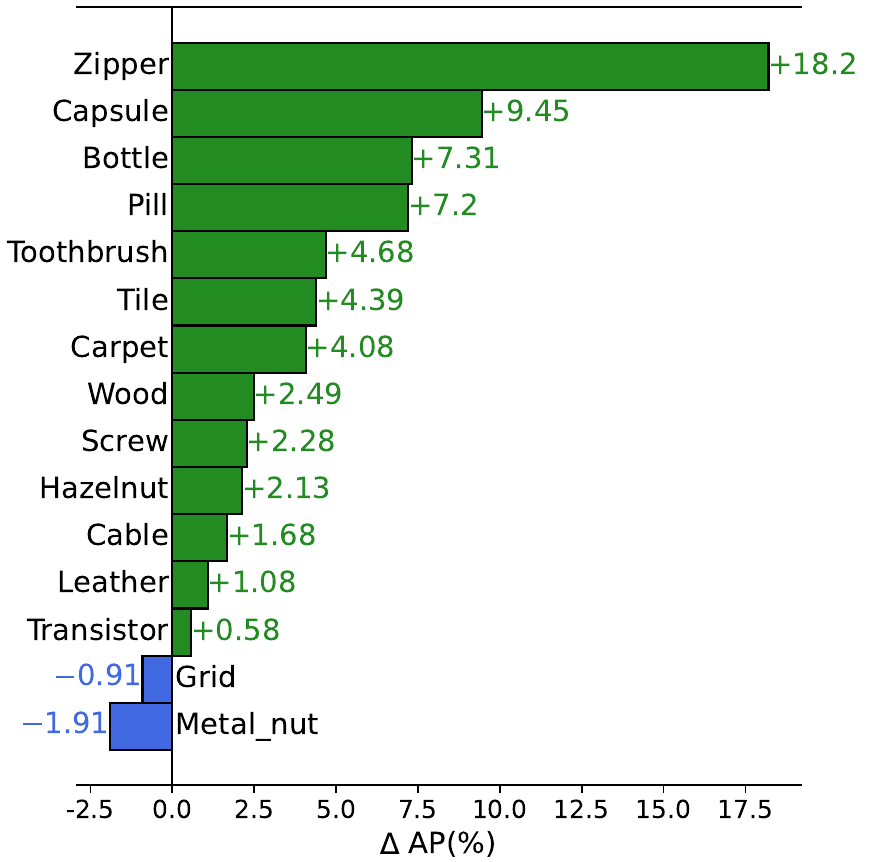}
				\caption{}
			\end{subfigure}%
			\begin{subfigure}[b]{0.5\textwidth}
				\centering
				\includegraphics[width=0.98\textwidth]{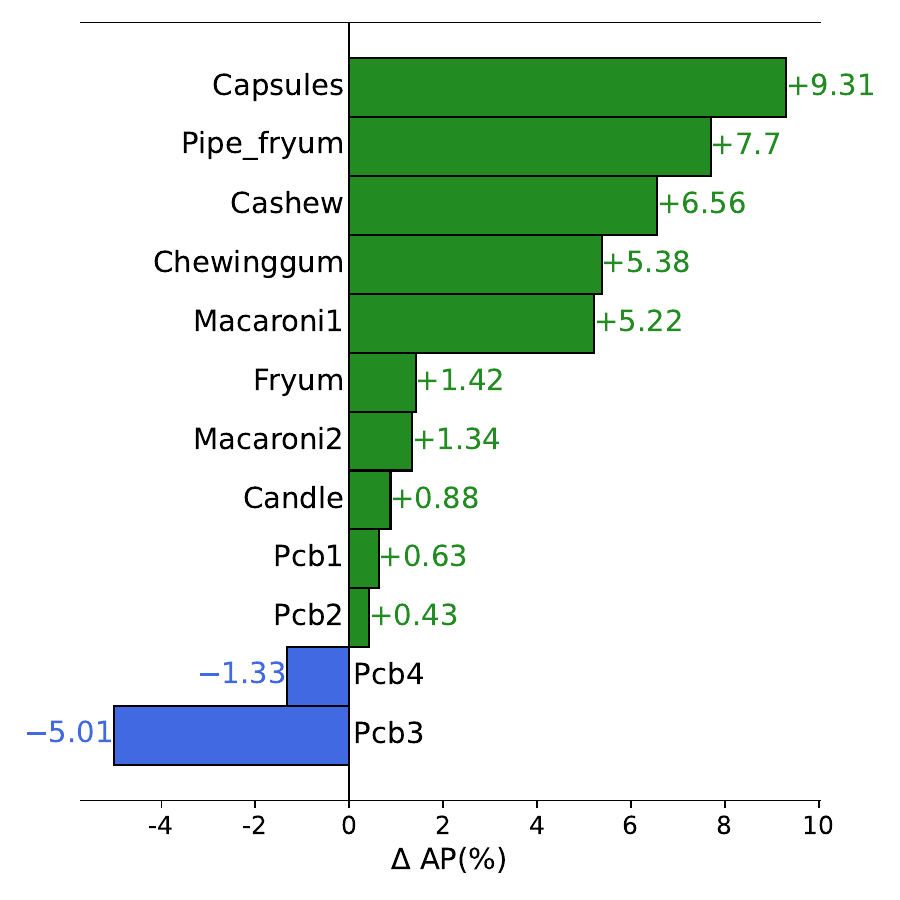}
				\caption{}
			\end{subfigure}
			\caption{The AP improvement of VCP-CLIP over the baseline for each product. (a) MVTec-AD (b) VisA}
			\label{fig6}
		\end{minipage}
		\hfill
		\begin{minipage}[b]{0.35\textwidth}
			\centering
			\includegraphics[width=0.9\textwidth]{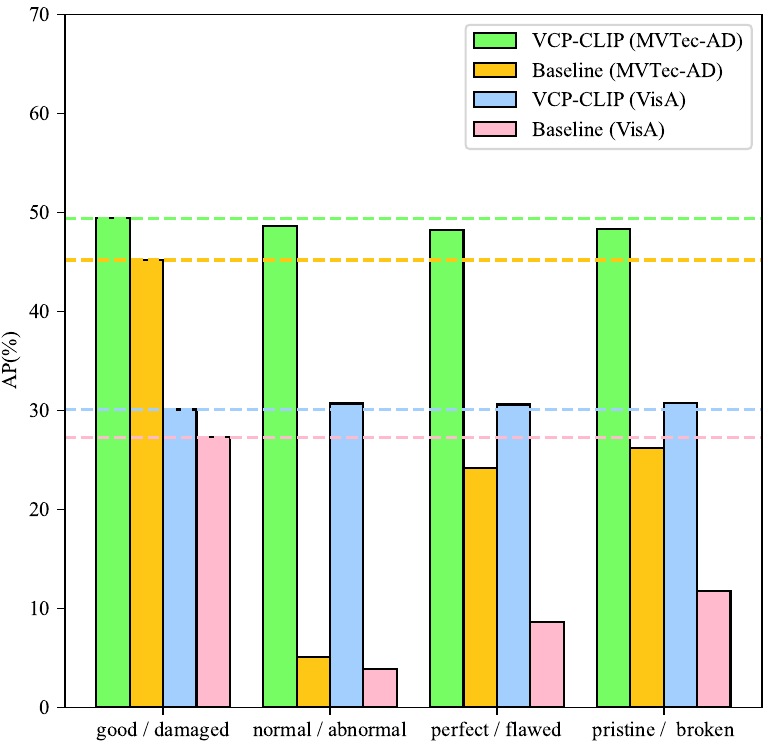}
			\caption{Performance comparison for different prompts during training and testing.}
			\label{fig7}
		\end{minipage}
		%\caption{Main Caption}
	\end{figure}

	\textbf{Same prompts during training and testing.} To better validate the effectiveness of VCP-CLIP, we compared it with the proposed baseline on MVTec-AD and VisA. Fig. \ref{fig6} illustrates the AP improvement of VCP-CLIP over the baseline for each product. In specific, VCP-CLIP demonstrates varying degrees of improvement among 13 out of the 15 products and 10 out of the 12 products on the MVTec-AD and VisA datasets, respectively. This affirms the robust generalization capability of VCP-CLIP, which is attributed to both the global visual context in Pre-VCP and the fine-grained local visual context in Post-VCP. 
	\par 
	\textbf{Different prompts during training and testing.} To validate the robustness of VCP-CLIP during the test process with different text prompts, we employed text prompts different from those used during training on the MVTec-AD and VisA datasets. Specifically, during training, the default state words "good/damaged" were used. During testing, we reported the metric AP when the state words were respectively "normal/abnormal", "perfect/flawed", and "pristine/broken". As shown in Fig. \ref{fig7}, our baseline performance sharply declined on two datasets, while the performance of VCP-CLIP remained relatively stable. This indicates that after incorporating VCP, the model can adaptively adjust the output text embeddings based on input images, thereby avoiding dependence on the specific text prompts used during training.
	
	\label{sec:blind}
	
	\subsection{Ablation Studies}
	% \usepackage{multirow}
	\begin{table}[tb]
		\centering
		\begin{minipage}[t]{0.45\textwidth}
			\centering
			\caption{Ablation on different components.}
			\label{tab2}
			\renewcommand{\arraystretch}{1.3}
			\resizebox{1.1\columnwidth}{!}
			{
				\begin{tabular}{*{6}{>{\centering\arraybackslash}p{1.2cm}}}
					\toprule
					\multirow{2}{*}{DTP} & \multicolumn{2}{c}{VCP} & \multirow{2}{*}{AUROC} & \multirow{2}{*}{PRO} & \multirow{2}{*}{AP} \\
					& Pre    & Post   &                        &                      &                     \\  \midrule
					$\surd$                      &            & $\surd$           & 91.4                   & 86.6                 & 47.5                \\
					$\surd$                      & $\surd$           &            & 90.4                   & 86.0                 & 46.1                \\
					& $\surd$           & $\surd$           & \underline{91.7}                   & \underline{86.7}                 & \underline{48.2 }               \\
					$\surd$                      &            &            & 89.2                   & 85.8                 & 45.2                \\
					$\surd$                      & $\surd$           & $\surd$           & \textbf{92.0}                   & \textbf{87.3}                 & \textbf{49.4}        \\  \bottomrule        
				\end{tabular}
			}
		\end{minipage}\hfill
		\begin{minipage}[t]{0.45\textwidth}
			\centering
			\caption{Ablation on ensemble of different patch-level image layers.}
			\label{tab3}
			\resizebox{1\columnwidth}{!}
			{
				\begin{tabular}{>{\centering\arraybackslash}p{2.2cm}*{6}{>{\centering\arraybackslash}p{1.2cm}}}
					\toprule
					Image layers & AUROC & PRO  & AP   \\  \midrule
					\{6\}                      & 79.6  & 65.6 & 22.5 \\
					\{12\}                     & 91.4  & 84.8 & 44.1 \\
					\{18\}                     & 91.2  & 84.4 & 44.5 \\
					\{24\}                     & 90.1  & 80.2 & 38.2 \\
					\{6, 12\}                  & 91.1  & 85.9 & 46.2 \\
					\{6, 12, 18\}              & \underline{91.8}  & \underline{87.1} & \underline{49.2} \\
					\{6, 12, 18, 24\}          & \textbf{92.0}  & \textbf{87.3} & \textbf{49.4}   \\   \bottomrule    
				\end{tabular}
			}
		\end{minipage}
	\end{table}
	\begin{table}[tb]
		\caption{Ablation on different template and state words in text prompts.}
		\label{tab4}
		\centering
		\resizebox{0.7\columnwidth}{!}
		{
			\begin{tabular}{>{\centering\arraybackslash}p{3cm}>{\centering\arraybackslash}p{3cm}*{3}{>{\centering\arraybackslash}p{1.2cm}}}
				\toprule
				Template	& State words          & AUROC & PRO  & AP   \\    \midrule
				\multirow{5}{*} {\makecell[c]{this is a [state]\\ photo of [$z(x,v)$]}} & perfect /  flawed    & 91.9  & \textbf{87.3} & 48.5 \\
				& normal / abnormal    & 90.1  & 86.1 & 48.7 \\
				& flawless / imperfect & 91.3  & 87.0 & 49.0 \\
				& pristine / broken    & 91.2  & 86.5 & 48.8 \\
				& good / damaged       & 91.5  & 86.9 & 49.1 \\
				\multirow{5}{*}{\makecell[c]{a photo of \\ a  [state] [$z(x,v)$]}}  & perfect /  flawed    & 91.8  & \underline{87.2} & 48.8 \\
				& normal / abnormal    & 91.7  & 86.6 & 48.7 \\
				& flawless / imperfect & \textbf{92.1}  & \underline{87.2} & 49.2 \\
				& pristine / broken    & \underline{92.0}  & 87.1 & \underline{49.3} \\
				& good / damaged       & \underline{92.0}  & \textbf{87.3} & \textbf{49.4}   \\   \bottomrule    
			\end{tabular}
		}
	\end{table}
	
	\begin{table}[t]
		\centering
		\begin{minipage}[t]{0.47\textwidth}
			\centering
			\caption{Ablation on different input resolutions upon VCP-CLIP.}
			\label{tab5}
			\renewcommand{\arraystretch}{1.2}
			\resizebox{1\columnwidth}{!}
			{
				\begin{tabular}{>{\centering\arraybackslash}p{3cm}*{3}{>{\centering\arraybackslash}p{1.2cm}}>{\centering\arraybackslash}p{1.5cm}}
					\toprule
					Input   resolution & AUROC & PRO  & AP   & Time (ms) \\  \midrule
					$224^2$                & 91    & 84.6 & 38.1 & \textbf{101.6}     \\
					$336^2$                & 90.7  & 87.6 & 44.9 & 104.5     \\
					$518^2$                & \textbf{92.0}  & \textbf{87.3} & \textbf{49.4} & 127.9     \\
					$546^2$                & 91.2  & 85.3 & 45.1 & 134.9     \\
					$798^2$                & 90.8  & 85   & 38.4 & 265.3  \\  \bottomrule    
				\end{tabular}
			}
		\end{minipage}\hfill
		\begin{minipage}[t]{0.47\textwidth}
			\centering
			\caption{Ablation on different Pre-trained backbone upon VCP-CLIP.}
			\label{tab6}
			\renewcommand{\arraystretch}{1.7}
			\resizebox{1\columnwidth}{!}
			{
				\begin{tabular}{>{\centering\arraybackslash}p{3cm}*{3}{>{\centering\arraybackslash}p{1.2cm}}>{\centering\arraybackslash}p{1.5cm}}
					\toprule
					Pretrained   backbone & AUROC & PRO  & AP   & Time (ms) \\   \midrule
					ViT-B-16-224          & 89.4  & 82.2 & 37.9 & \textbf{84.4}      \\
					ViT-L-14-224          & 91.9  & 85.7 & 43.3 & 105.1     \\
					ViT-L-14-336          & \textbf{92.0}  & \textbf{87.3} & \textbf{49.4} & 127.9    \\ \bottomrule    
				\end{tabular}
			}
		\end{minipage}
	\end{table}
	\textbf{Influence of different components.} To assess the impact of different components on VCP, experiments were conducted on MVTec-AD. Results in Table \ref{tab2} indicate performance when using DTP, Pre-VCP or Post-VCP individually. Notably, the optimal performance for VCP is achieved when all combined. It can been seen that the performance decline is more pronounced after removing Post-VCP compared to Pre-VCP. We also attempted to remove the learnable text embeddings from each layer of the text encoder (without DTP), which resulted in a decrease of 0.3\% in AUROC, 0.6\% in PRO, and 1.2\% in AP. This is because the original text space cannot directly comprehend the global features of images, while DTP ensures deep fine-tuning of each text encoder layer, thereby fostering mutual understanding and fusion of different modalities.
	\par
	\textbf{Influence of ensemble of different patch-level image layers.} In Table \ref{tab3}, we explore the impact of patch-level features from different image encoder layers on VCP-CLIP’s performance. The experiments were conducted on the MVTec-AD dataset. An intuitive observation is that image features from intermediate layers (i.e. the 12th and 18th layers), contribute more to the final segmentation result. Image features from lower layers (i.e., the 6th layer) are too low-level, while those from higher layers (i.e., the 24th layer) are overly abstract. Their effectiveness is not as pronounced as those from intermediate layers. However, We observed a positive correlation between incorporated layer numbers and improved segmentation results. To maintain high performance, we adopted all patch-level features from \{6, 12, 18, 24\} layers in VCP-CLIP.
	\par 
	\textbf{Ablation on text prompt design.} As demonstrated in Table \ref{tab4}, we considered two commonly used text prompt templates and explored the impact of different prompting state words in the proposed VCP-CLIP on MVTec-AD. Specifically, we designed the following two text prompt templates: 1) this is a [state] photo of [$z(x,v)$]; 2) a photo of a [state] [$z(x,v)$]. The state words (e.g. "perfect/flawed") are respectively inserted into the template to generate normal and abnormal text prompts. It can be observed that for the same template with different state words, our VCP-CLIP model consistently maintains similar performance, validating the robustness towards the state words. Furthermore, the second type of template, default employed in VCP-CLIP, outperforms the first type overall, which may be attributed to the repeated usage of similar template during the pre-training process of the vanilla CLIP.
	\par
	\textbf{Ablation on different pretrained models and resolutions.} In Table \ref{tab5} and Table \ref{tab6}, we conducted a comprehensive analysis of the impact of varying input image resolution and pre-trained backbone on MVTec-AD. The former is tested using ViT-L-14-336, while the latter reports the optimal performance under different backbones pre-trained by OpenAI. The inference time was simultaneously tested for a single image (average of 200 images). We observe that a moderate increase in input image resolution contributes to more precise segmentation (higher AP). However, deviations from the original pre-training resolution ($336^2$ to $798^2$), leading to model degradation. This outcome can be attributed to the model deviating from the original image space. The result in Table \ref{tab6} shows that our VCP-CLIP achieves the optimal segmentation performance in ViT-L-14-336. Therefore, we have chosen it as the default backbone.
	\section{Conclusion}
	In this paper, we present VCP-CLIP, a novel zero-shot anomaly segmentation (ZSAS) method achieved through the integration of visual context  prompting (VCP). The core methodology involves incorporating richer visual knowledge into the textual space and cross-modal interaction between textual and visual features. Specifically, a Pre-VCP and a Post-VCP module are designed to respectively introduce global and fine-grained image features into the textual space. With this design, our model can directly segment anomalies in novel products without any prior knowledge. Extensive experiments conducted on 10 real-world industrial anomaly segmentation datasets showcase VCP-CLIP’s state-of-the-art performance in ZSAS. 
	\par 
	\textbf{Acknowledgments.} This work was supported in part by the National Science and Technology Major Project of China under Grant 2022ZD0119402; in part  by the National Natural Science Foundation of China under Grant No. 62373350 and U21A20482; in part  by the Youth Innovation Promotion Association CAS (2023145) ; in part  by the Beijing Municipal Natural Science Foundation, China, under Grant L243018.

\bibliographystyle{splncs04}
\bibliography{main}

\begin{thebibliography}{10}
\providecommand{\url}[1]{\texttt{#1}}
\providecommand{\urlprefix}{URL }
\providecommand{\doi}[1]{https://doi.org/#1}

\bibitem{LayerNorm}
Ba, J.L., Kiros, J.R., Hinton, G.E.: Layer normalization. arXiv preprint
  arXiv:1607.06450  (2016)

\bibitem{MVTec}
Bergmann, P., Fauser, M., Sattlegger, D., Steger, C.: Mvtec ad--a comprehensive
  real-world dataset for unsupervised anomaly detection. In: Proceedings of the
  IEEE/CVF conference on computer vision and pattern recognition. pp.
  9592--9600 (2019)

\bibitem{KSDD2}
Bo{\v{z}}i{\v{c}}, J., Tabernik, D., Sko{\v{c}}aj, D.: Mixed supervision for
  surface-defect detection: From weakly to fully supervised learning. Computers
  in Industry  \textbf{129},  103459 (2021)

\bibitem{SAA}
Cao, Y., Xu, X., Sun, C., Cheng, Y., Du, Z., Gao, L., Shen, W.: Segment any
  anomaly without training via hybrid prompt regularization. arXiv preprint
  arXiv:2305.10724  (2023)

\bibitem{VAND}
Chen, X., Han, Y., Zhang, J.: A zero-/few-shot anomaly classification and
  segmentation method for cvpr 2023 vand workshop challenge tracks 1\&2: 1st
  place on zero-shot ad and 4th place on few-shot ad. arXiv preprint
  arXiv:2305.17382  (2023)

\bibitem{CLIPAD}
Chen, X., Zhang, J., Tian, G., He, H., Zhang, W., Wang, Y., Wang, C., Wu, Y.,
  Liu, Y.: Clip-ad: A language-guided staged dual-path model for zero-shot
  anomaly detection. arXiv preprint arXiv:2311.00453  (2023)

\bibitem{vicuna}
Chiang, W.L., Li, Z., Lin, Z., Sheng, Y., Wu, Z., Zhang, H., Zheng, L., Zhuang,
  S., Zhuang, Y., Gonzalez, J.E., Stoica, I., Xing, E.P.: Vicuna: An
  open-source chatbot impressing gpt-4 with 90\%* chatgpt quality (2023),
  \url{https://lmsys.org/blog/2023-03-30-vicuna/}

\bibitem{AnVoL}
Deng, H., Zhang, Z., Bao, J., Li, X.: Anovl: Adapting vision-language models
  for unified zero-shot anomaly localization. arXiv preprint arXiv:2308.15939
  (2023)

\bibitem{AnomalyGPT}
Gu, Z., Zhu, B., Zhu, G., Chen, Y., Tang, M., Wang, J.: Anomalygpt: Detecting
  industrial anomalies using large vision-language models. In: Proceedings of
  the AAAI Conference on Artificial Intelligence. vol.~38, pp. 1932--1940
  (2024)

\bibitem{WinCLIP}
Jeong, J., Zou, Y., Kim, T., Zhang, D., Ravichandran, A., Dabeer, O.: Winclip:
  Zero-/few-shot anomaly classification and segmentation. In: Proceedings of
  the IEEE/CVF Conference on Computer Vision and Pattern Recognition. pp.
  19606--19616 (2023)

\bibitem{VPT}
Jia, M., Tang, L., Chen, B.C., Cardie, C., Belongie, S., Hariharan, B., Lim,
  S.N.: Visual prompt tuning. In: European Conference on Computer Vision. pp.
  709--727. Springer (2022)

\bibitem{Adam}
Kingma, D.P., Ba, J.: Adam: A method for stochastic optimization. arXiv
  preprint arXiv:1412.6980  (2014)

\bibitem{SAM}
Kirillov, A., Mintun, E., Ravi, N., Mao, H., Rolland, C., Gustafson, L., Xiao,
  T., Whitehead, S., Berg, A.C., Lo, W.Y., et~al.: Segment anything. In:
  Proceedings of the IEEE/CVF International Conference on Computer Vision. pp.
  4015--4026 (2023)

\bibitem{CLIPSAM}
Li, S., Cao, J., Ye, P., Ding, Y., Tu, C., Chen, T.: Clipsam: Clip and sam
  collaboration for zero-shot anomaly segmentation. arXiv preprint
  arXiv:2401.12665  (2024)

\bibitem{focal}
Lin, T.Y., Goyal, P., Girshick, R., He, K., Doll{\'a}r, P.: Focal loss for
  dense object detection. In: Proceedings of the IEEE international conference
  on computer vision. pp. 2980--2988 (2017)

\bibitem{grounding}
Liu, S., Zeng, Z., Ren, T., Li, F., Zhang, H., Yang, J., Li, C., Yang, J., Su,
  H., Zhu, J., et~al.: Grounding dino: Marrying dino with grounded pre-training
  for open-set object detection. arXiv preprint arXiv:2303.05499  (2023)

\bibitem{GC}
Lv, X., Duan, F., Jiang, J.j., Fu, X., Gan, L.: Deep metallic surface defect
  detection: The new benchmark and detection network. Sensors  \textbf{20}(6),
  ~1562 (2020)

\bibitem{dice}
Milletari, F., Navab, N., Ahmadi, S.A.: V-net: Fully convolutional neural
  networks for volumetric medical image segmentation. In: 2016 fourth
  international conference on 3D vision (3DV). pp. 565--571. Ieee (2016)

\bibitem{BTech}
Mishra, P., Verk, R., Fornasier, D., Piciarelli, C., Foresti, G.L.: Vt-adl: A
  vision transformer network for image anomaly detection and localization. In:
  2021 IEEE 30th International Symposium on Industrial Electronics (ISIE). pp.
  01--06. IEEE (2021)

\bibitem{DAGM}
f{\"u}r Mustererkennung, D.A.: Weakly supervised learning for industrial
  optical inspection (2007)

\bibitem{GPT}
Ouyang, L., Wu, J., Jiang, X., Almeida, D., Wainwright, C., Mishkin, P., Zhang,
  C., Agarwal, S., Slama, K., Ray, A., et~al.: Training language models to
  follow instructions with human feedback. Advances in Neural Information
  Processing Systems  \textbf{35},  27730--27744 (2022)

\bibitem{CLIP}
Radford, A., Kim, J.W., Hallacy, C., Ramesh, A., Goh, G., Agarwal, S., Sastry,
  G., Askell, A., Mishkin, P., Clark, J., et~al.: Learning transferable visual
  models from natural language supervision. In: International conference on
  machine learning. pp. 8748--8763. PMLR (2021)

\bibitem{CoOp}
Radford, A., Kim, J.W., Hallacy, C., Ramesh, A., Goh, G., Agarwal, S., Sastry,
  G., Askell, A., Mishkin, P., Clark, J., et~al.: Learning transferable visual
  models from natural language supervision. In: International conference on
  machine learning. pp. 8748--8763. PMLR (2021)

\bibitem{DenseCLIP}
Rao, Y., Zhao, W., Chen, G., Tang, Y., Zhu, Z., Huang, G., Zhou, J., Lu, J.:
  Denseclip: Language-guided dense prediction with context-aware prompting. In:
  Proceedings of the IEEE/CVF Conference on Computer Vision and Pattern
  Recognition. pp. 18082--18091 (2022)

\bibitem{BSD}
Schlagenhauf, T., Landwehr, M.: Industrial machine tool component surface
  defect dataset. Data in Brief  \textbf{39},  107643 (2021)

\bibitem{Road}
Shi, Y., Cui, L., Qi, Z., Meng, F., Chen, Z.: Automatic road crack detection
  using random structured forests. IEEE Transactions on Intelligent
  Transportation Systems  \textbf{17}(12),  3434--3445 (2016)

\bibitem{tao}
Tao, X., Gong, X., Zhang, X., Yan, S., Adak, C.: Deep learning for unsupervised
  anomaly localization in industrial images: A survey. IEEE Transactions on
  Instrumentation and Measurement  (2022)

\bibitem{llama}
Touvron, H., Lavril, T., Izacard, G., Martinet, X., Lachaux, M.A., Lacroix, T.,
  Rozi{\`e}re, B., Goyal, N., Hambro, E., Azhar, F., et~al.: Llama: Open and
  efficient foundation language models. arXiv preprint arXiv:2302.13971  (2023)

\bibitem{RSDD}
Yu, H., Li, Q., Tan, Y., Gan, J., Wang, J., Geng, Y.a., Jia, L.: A
  coarse-to-fine model for rail surface defect detection. IEEE Transactions on
  Instrumentation and Measurement  \textbf{68}(3),  656--666 (2018)

\bibitem{mobileSAM}
Zhang, C., Han, D., Qiao, Y., Kim, J.U., Bae, S.H., Lee, S., Hong, C.S.: Faster
  segment anything: Towards lightweight sam for mobile applications. arXiv
  preprint arXiv:2306.14289  (2023)

\bibitem{MSD}
Zhang, J., Ding, R., Ban, M., Guo, T.: Fdsnet: An accurate real-time surface
  defect segmentation network. In: ICASSP 2022-2022 IEEE International
  Conference on Acoustics, Speech and Signal Processing (ICASSP). pp.
  3803--3807. IEEE (2022)

\bibitem{CoCoOp}
Zhou, K., Yang, J., Loy, C.C., Liu, Z.: Conditional prompt learning for
  vision-language models. In: Proceedings of the IEEE/CVF Conference on
  Computer Vision and Pattern Recognition. pp. 16816--16825 (2022)

\bibitem{AnomalyCLIP}
Zhou, Q., Pang, G., Tian, Y., He, S., Chen, J.: Anomalyclip: Object-agnostic
  prompt learning for zero-shot anomaly detection. In: The Twelfth
  International Conference on Learning Representations (2023)

\bibitem{ZigCLIP}
Zhou, Z., Lei, Y., Zhang, B., Liu, L., Liu, Y.: Zegclip: Towards adapting clip
  for zero-shot semantic segmentation. In: Proceedings of the IEEE/CVF
  Conference on Computer Vision and Pattern Recognition. pp. 11175--11185
  (2023)

\bibitem{VisA}
Zou, Y., Jeong, J., Pemula, L., Zhang, D., Dabeer, O.: Spot-the-difference
  self-supervised pre-training for anomaly detection and segmentation. In:
  European Conference on Computer Vision. pp. 392--408. Springer (2022)

\end{thebibliography}
\newpage
\appendix
\begin{center}
	\large \textbf{Appendix}
\end{center}
\par
This supplementary appendix contains the following five parts: 1) Detailed experimental setup and introduction of state-of-the-art methods in Section \ref{secA}; 2) Additional experiments and further analysis in Section \ref{secB}; 3) Introduction of the real industrial datasets in Section \ref{secC}; 4) More detailed presentations of quantitative and qualitative results in Section \ref{secD}; 5) Discussion of model limitations in Section \ref{secE}.  
\section{Implementation details and state-of-the-art methods} \label{secA}

\subsection{Implementation details}

\textbf{Details of model configuration.} In this paper, we adopt the CLIP model with ViT-L-14-336 pretrained by OpenAI \cite{CLIP}. The length of learnable category vectors (tokens) $r$ and the length of learnable text embeddings $n$ in each text encoder layer are set to 2 and 1, respectively. The number of image encoder layers $B$  for extracting patch-level features is set to 4 and we evenly select the 6th, 12th, 18th, and 24th layers to acquire dense image embeddings. In our baseline, we use a single linear layer to map the dense image embeddings $Z_s^l$  to  $F_s^l$ in the joint embedding space. In addition, the number of attention heads $M$ in Post-VCP module is set to 8 and the fusion weight $\alpha$  is set to 0.75 in our VCP-CLIP. All images are resized to a resolution of 518 × 518, and then fed into the image encoder.
\par
\textbf{Details of training and testing.} We conduct experiments on 10 publicly available real-world industrial anomaly segmentation datasets, including MVTec-AD and VisA, which are widely used in previous ZSAS tasks. \textbf{Notably, the products in VisA do not overlap with those in other datasets. Therefore, to evaluate ZSAS performance on other datasets, we employ weights trained on VisA's test sets. As for VisA, we assess ZSAS performance after training on MVTec-AD.} The final results for each dataset are derived from the average value of the products it contains. During training on seen products, we maintain the original CLIP parameters in a frozen state, updating only the newly introduced learnable parameters. The Adam optimizer \cite{Adam} with an initial learning rate of 4e-5 is used and the model is trained for continuous 10 epochs with a batch size of 32. All experiments are conducted on a single NVIDIA GeForce RTX 3090, and we perform three runs using different random seeds and then average the results. 
\subsection{State-of-the-art methods}

\begin{itemize}[label=$\diamond$]
	\item \textbf{WinCLIP} \cite{WinCLIP} is a representative ZSAS method that ensembles a large number of manually designed text prompts to classify sub-images within each window. The classification outcomes from various scaled windows are then aggregated to derive the ultimate anomaly segmentation results. The values of AUROC and PRO on MVTec-AD and VisA are obtained from the original paper, while the results for other metrics and datasets are based on our code reproduction following the settings specified in the original paper. 
	\item \textbf{AnVoL} \cite{AnVoL} optimizes model parameters to adapt to ZSAS tasks by utilizing a test-time adaptation technique. The values of AUROC and PRO on MVTec-AD and VisA are obtained from the original paper, while other metrics and datasets are derived from the implementation of the official code.
	\item \textbf{CoCoOp} \cite{CoCoOp} is a method that applies CLIP to image classification tasks based on prompt learning. It uses continuously learnable vectors instead of manually designed text prompts, enhancing the model's generalization to novel classes by making the prompt conditioned on each input image. To adapt CoCoOp to ZSAS task, we improve the prompt templates used in the original paper. In specific, the original template  $[v_1(x)][v_2(x)]\cdots[v_r(x)]\allowbreak[class]$ is replaced with $[v_1(x)][v_2(x)]\cdots[v_r(x)][good][class]$ and $[v_1(x)][v_2(x)]\allowbreak\cdots[v_r(x)][damaged][class]$ for the generation of normal and abnormal text prompts, where $v_i(x)$  represents the learnable word embeddings that incorporate image features $x$. The anomaly segmentation results are obtained in the same manner as our baseline, with all other parameters remaining unchanged from the original paper.
	\item \textbf{AnomalyGPT} \cite{AnomalyGPT} integrates a large language model for anomaly segmentation and supports multi-turn dialogues with users. It employs supervised training using synthetic anomaly data to enable the model to generalize to new products. Additionally, it supports fine-tuning the model using finely annotated data to achieve better ZSAS performance. For product category descriptions in MVTec-AD and VisA, we adhere to the original settings. For other datasets, we use the following product description: "\textit{This is a photo of a [class] for anomaly detection, which should be without any damage, flaw, defect, scratch, hole, or broken part}". We conducted experiments using official code and evaluated the model's ZSAS performance in the same manner as VCP-CLIP.
	\item \textbf{APRIL-GAN} \cite{VAND} adopts a text prompting design strategy similar to WinCLIP, and it fine-tunes the model using auxiliary datasets to adapt to the ZSAS task. We conducted experiments using official code and pretrained weights, keeping all parameters and settings consistent with the original paper. 
	\item \textbf{CLIP-AD} \cite{CLIPAD} utilizes a text prompt design similar to WinCLIP and adapts to the ZSAS task through feature surgery and fine-tuning techniques. Due to the official code not being open-sourced, it is only compared as a concurrent work with our VCP-CLIP on MVTec-AD and VisA.
	\item \textbf{ClipSAM} \cite{CLIPSAM} is a collaboration framework of CLIP and SAM, which are respectively used for rough and fine segmentation of abnormal regions. In this paper, it is compared as a concurrent work with our VCP-CLIP on MVTec-AD and VisA.
	\item \textbf{AnomalyCLIP} \cite{AnomalyCLIP} proposes to learn object-agnostic text prompts for ZSAS. It uses [object] to replace specific product categories [class] in the text prompts, thereby focusing the model on the abnormal regions of images. Due to the official code not being open-sourced, it is compared as a concurrent work with our VCP-CLIP on MVTec-AD and VisA.
\end{itemize}
\section{Additional results and ablations}  \label{secB}
\subsection{Comparison with concurrent methods}
In addition to the state-of-the-art methods that have already been compared, we also pay attention to three other concurrent works for ZSAS, namely CLIP-AD \cite{CLIPAD}, ClipSAM \cite{CLIPSAM} and AnomalyCLIP \cite{AnomalyCLIP}. They all utilize auxiliary datasets to fine-tune foundation models such as CLIP and SAM, to adapt to the ZSAS task. Table \ref{tab7} presents a quantitative comparison between the other three methods and our VCP-CLIP. Given that these methods utilize the same experimental setup as ours, all experimental results are directly sourced from their respective original papers.
\par
Our method demonstrates superior performance compared to the others in terms of AUROC, PRO, and AP on the VisA dataset, showcasing remarkable results. On the MVTec-AD dataset, our VCP-CLIP also achieves comparable ZSAS performance. Despite potentially having lower AUROC and PRO scores compared to ClipSAM, VCP-CLIP exhibits a higher AP, indicating more precise segmentation results. Furthermore, ClipSAM combines two foundation models, CLIP \cite{CLIP} and SAM \cite{SAM}, which may reduce model inference efficiency and render it less suitable for real-world industrial applications.

\begin{table}[t]
	\caption{Comparison with concurrent ZSAS methods. The best results are shown in \textbf{bold} and the second best results are \underline{underlined}.}
	\label{tab7}
	\centering
	\renewcommand{\arraystretch}{1.2}
	\resizebox{0.7\columnwidth}{!}
	{
		\begin{tabular}{>{\centering\arraybackslash}p{2cm} *{6}{>{\centering\arraybackslash}p{1.4cm}}}
			\toprule
			\multirow{2}{*}{} & \multicolumn{3}{c}{MVTec-AD} & \multicolumn{3}{c}{VisA} \\  \cmidrule(lr){2-7}
			& AUROC    & PRO     & AP      & AUROC   & PRO    & AP    \\  \midrule
			CLIP-AD           & 91.2     & 85.6    & 39.4    & 95.2    & \underline{88.7}   & 24.2  \\
			ClipSAM           & \textbf{92.3}     & \textbf{88.3}    & \underline{45.9}    & \underline{95.6}    & 87.5   & 26.0    \\
			AnomalyCLIP       & 91.1     & 81.4    & --      & 95.5    & 87.0     & --    \\
			VCP-CLIP          &\underline{92.0}       & \underline{87.3}    & \textbf{49.4}    & \textbf{95.7}    & \textbf{90.0}    & \textbf{30.1}  \\ \bottomrule
		\end{tabular}
	}
\end{table}

\subsection{Additional ablations}
In this subsection, we explore the impact of the position of learnable vectors in deep text prompting (DTP) and the hyperparameters on VCP-CLIP.
\par
\begin{figure}[t]
	\centering
	\begin{subfigure}[b]{0.32\textwidth}
		\centering
		\includegraphics[width=\textwidth]{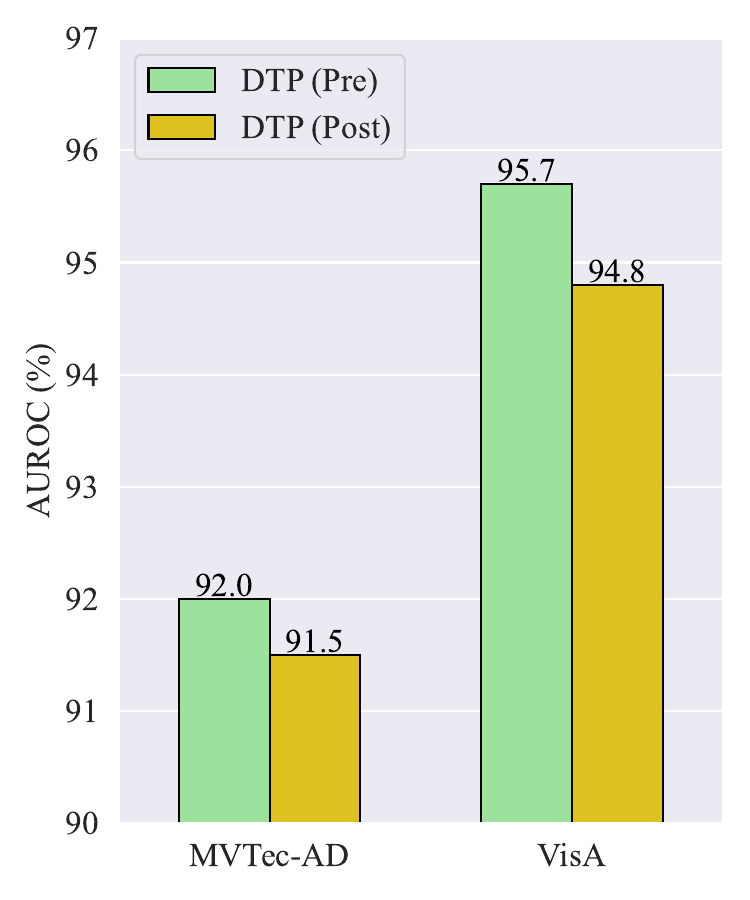}
		\caption{}
		\label{fig:sub1}
	\end{subfigure}
	\hfill
	\begin{subfigure}[b]{0.32\textwidth}
		\centering
		\includegraphics[width=\textwidth]{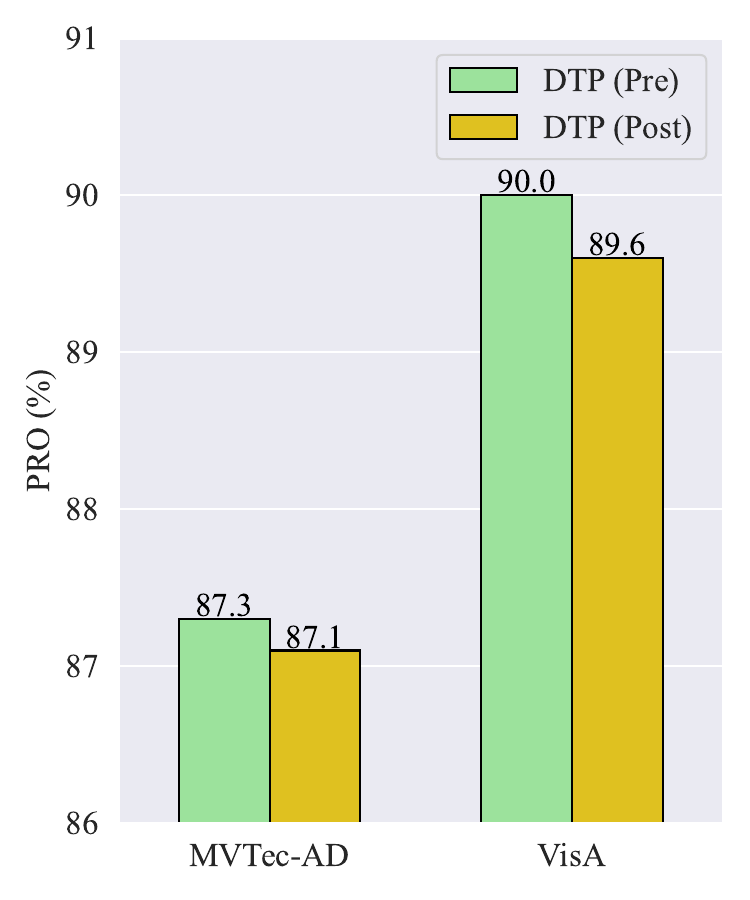}
		\caption{}
		\label{fig:sub2}
	\end{subfigure}
	\hfill
	\begin{subfigure}[b]{0.32\textwidth}
		\centering
		\includegraphics[width=\textwidth]{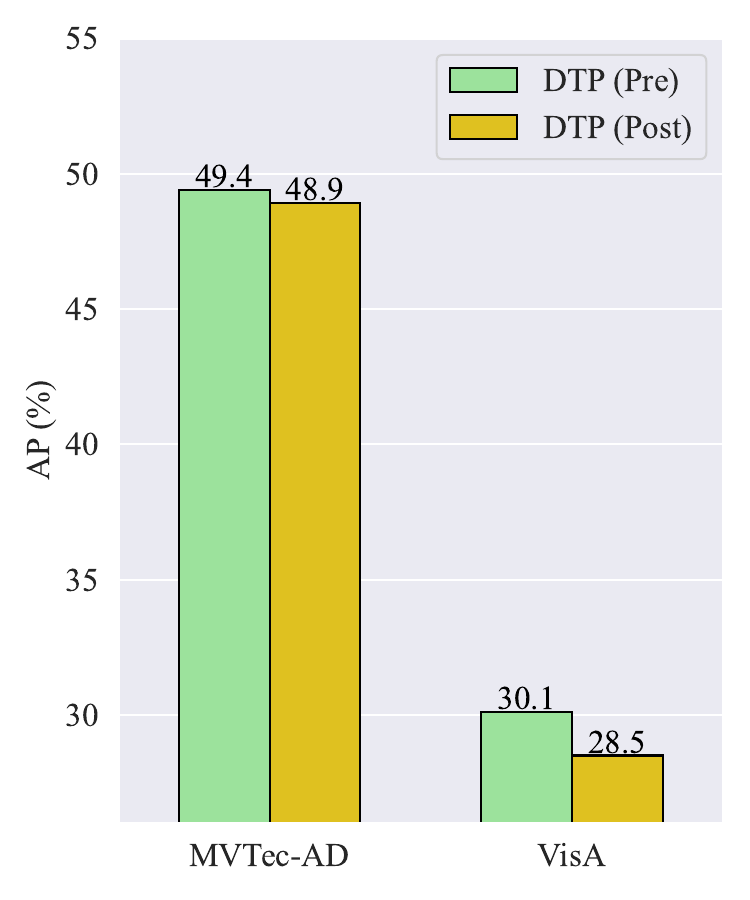}
		\caption{}
		\label{fig:sub3}
	\end{subfigure}
	\caption{Ablation on deep text prompting (DTP). We compare the learnable text embeddings when they are respectively at the beginning and end of the input prompts, denoted as DTP (Pre) and DTP (Post). (a) AUROC (b) PRO (c) AP}
	\label{fig8}
\end{figure}

 
\textbf{Ablation on DTP}. As mentioned in Section 3.2, due to the utilization of masked self-attention in the text encoder, the model does not attend to the context of future tokens in the text prompts. Consequently, the placement of learnable text embeddings at the start and end of a sentence produces different outcomes. In other words, $[s_i, P_i, H_i, e_i, J_i]$ and $[s_i, H_i, P_i, e_i, J_i]$ are not mathematically equivalent, where $[\cdot,\cdot]$ represents the concatenation operation on the sequence length dimension. We refer to these two scenarios as DTP (Pre) and DTP (Post) and compare the performance of ZSAS as depicted in Fig. \ref{fig8}. It is evident that the performance of DTP (Post) is slightly lower than that of DTP (Pre). This is because placing learnable text embeddings at the beginning of a sentence can influence the entire sentence, thereby benefiting the textual space refinement. 
\par 
\begin{figure}[t]
	\centering
	\begin{subfigure}[b]{0.49\textwidth}
		\centering
		\includegraphics[width=\textwidth]{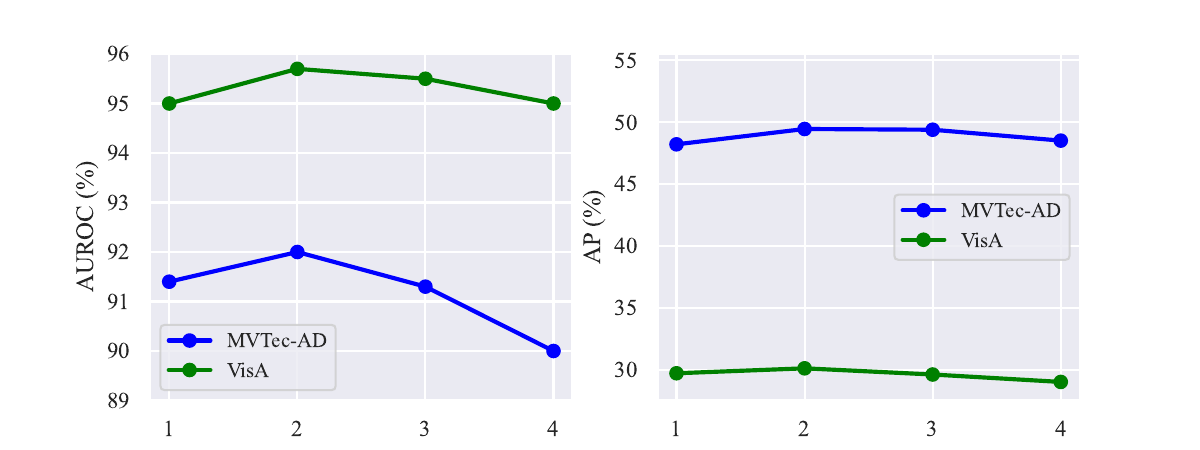}
		\caption{}
		\label{fig:sub1}
	\end{subfigure}
	\hfill
	\begin{subfigure}[b]{0.49\textwidth}
		\centering
		\includegraphics[width=\textwidth]{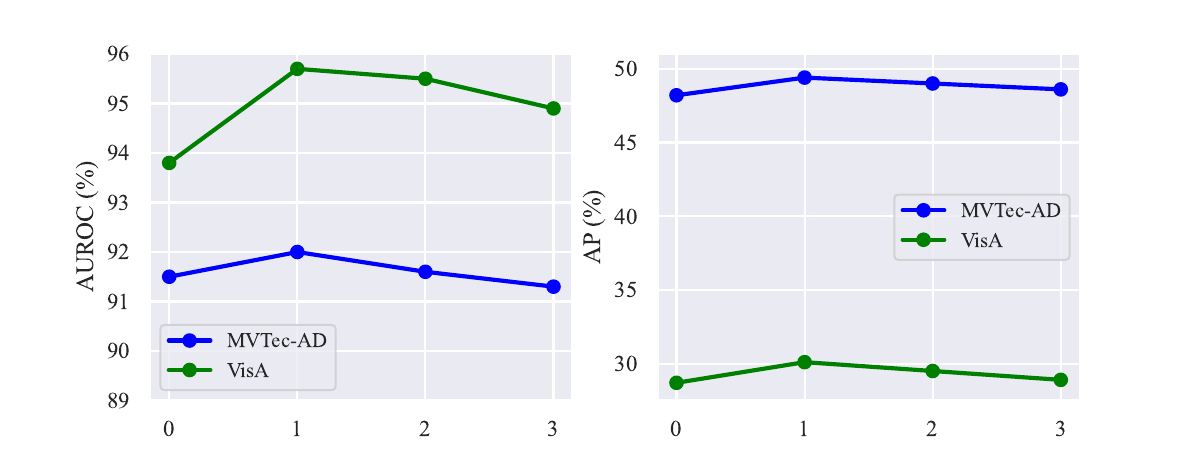}
		\caption{}
		\label{fig:sub2}
	\end{subfigure}
	\vfill
	\begin{subfigure}[b]{0.49\textwidth}
		\centering
		\includegraphics[width=\textwidth]{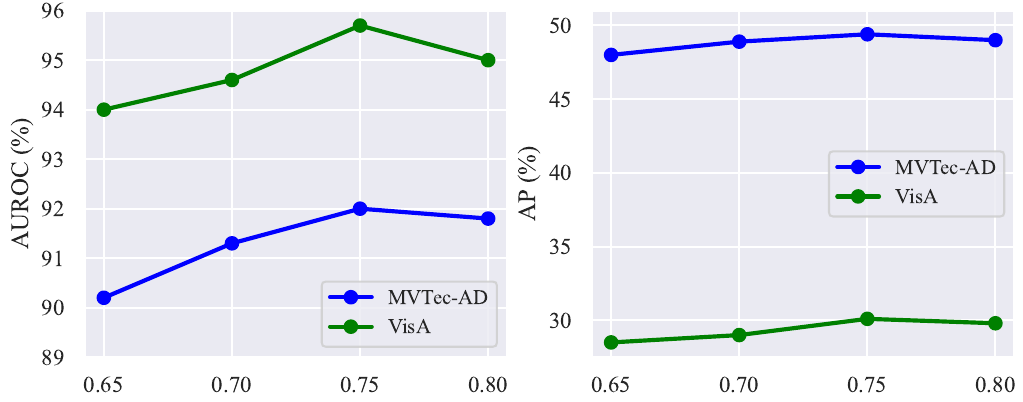}
		\caption{}
		\label{fig:sub2}
	\end{subfigure}
	\hfill
	\begin{subfigure}[b]{0.49\textwidth}
		\centering
		\includegraphics[width=\textwidth]{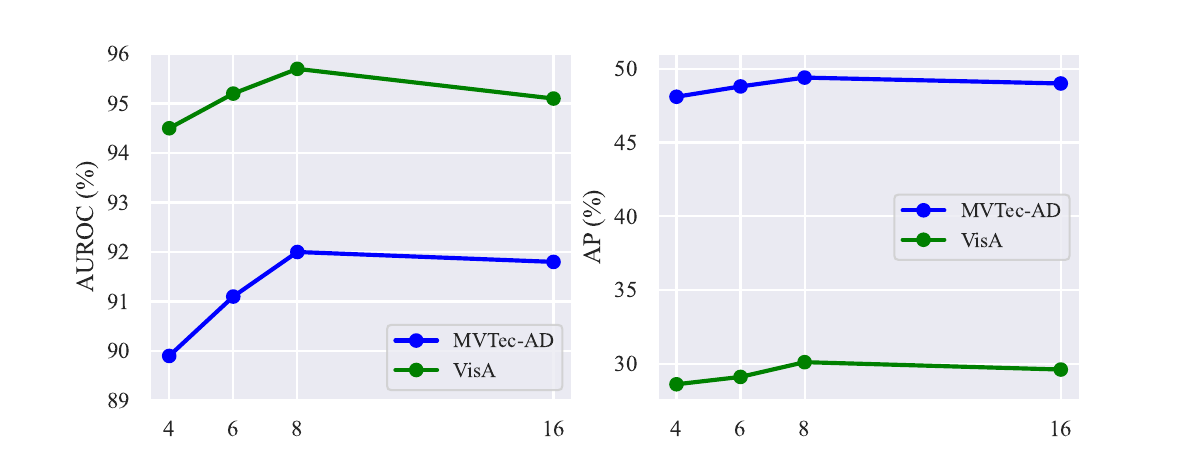}
		\caption{}
		\label{fig:sub2}
	\end{subfigure}
	\caption{(a) Ablation on the length of learnable category vectors $r$. (b) Ablation on the length of learnable text embeddings in each text encoder layer $n$. (c) Ablation on the fusion weight of different anomaly maps $\alpha$. (d) Ablation on the number of attention heads in the Post-VCP module $M$.}
	\label{fig9}
\end{figure}
\textbf{Ablation on hyperparameters}. As shown in Fig. \ref{fig9}, we explore the impact of different hyperparameters on the performance of VCP-CLIP, including the length of learnable category vectors $r$, the length of learnable text embeddings $n$ in each text encoder layer, the fusion weight of different anomaly maps $\alpha$, and the number of attention heads $M$ in the Post-VCP module.
\par 
1) Fig. \ref{fig9}(a) shows that the model performs best when the length of learnable category vectors $r$ is set to 2. Using too many or too few vectors hinders the model's segmentation performance. This aligns with our expectations, as typically, two tokens are adequate for representing product categories, while an excess of tokens may introduce unnecessary semantic overlap; 2) Fig. \ref{fig9}(b) illustrates that the model achieves the highest AUROC and AP when including one learnable text embedding $(n=1)$ in each layer of the text encoder. However, as $n$ increases, the model's performance starts to decline. This phenomenon occurs because excessive refinement of the textual space can diminish the original CLIP's generalization capability and may lead to overfitting on the limited training data; 3) The fusion weight $\alpha$ determines the relative importance of the anomaly maps $M_1$ from the baseline and $M_2$ from the additional VCP module. As depicted in Fig. \ref{fig9}(c), our model achieves superior fusion results with $\alpha$ set to 0.75. It's important to note that a higher $\alpha$ value indicates a greater contribution from the anomaly map $M_2$ generated by the VCP module, highlighting the effectiveness of using visual context prompting; 4) Fig. \ref{fig9}(d) outlines the impact of the number of attention heads $M$ in the Post-VCP module. To achieve optimal performance, we set $M = 8$ to leverage multiple attention heads for focusing on detailed image features and updating text embeddings accordingly.

\subsection{Additional analysis}

\begin{table}[t]
	\caption{Comparison of average inference time and maximum GPU cost on MVTec-AD. The best results are shown in \textbf{bold} and the second best results are \underline{underlined}.}
	\label{tab8}
	\centering
	\renewcommand{\arraystretch}{1.2}
	\resizebox{0.7\columnwidth}{!}
	{
		\begin{tabular}{>{\centering\arraybackslash}p{2.5cm} *{3}{>{\centering\arraybackslash}p{1.4cm}}*{2}{>{\centering\arraybackslash}p{2.4cm}}}
			\toprule
			& AUROC & PRO  & AP   & Time (ms) & GPU cost (GB) \\  \midrule
			WinCLIP         & 85.1  & 64.6 & 18.2 & 840.0       & \textbf{2.02}          \\
			AnVoL           & \underline{90.6}  & 77.8 & 28.1 & 132.1     & 3.27          \\
			CoCoOp          & 88.2  & 83.2 & 40.4 & \underline{108.2}     & \underline{3.01}          \\
			AnomalyGPT      & 79.5  & 45.9 & 23.7 & 2506.0      & 17.43         \\
			APRIL-GAN       & 87.6  & 44.0   & 40.8 & \textbf{105.0}       & 3.32          \\
			Baseline(ours) & 89.2  & \underline{85.8} & \underline{45.2} & 114.5     & 3.02          \\
			VCP-CLIP(ours) & \textbf{92.0}    & \textbf{87.3} & \textbf{49.4} & 127.9     & 3.05         \\  \bottomrule
		\end{tabular}
	}
\end{table}

\begin{figure}[t]
	\centering
	\begin{subfigure}[b]{0.51\textwidth}
		\centering
		\includegraphics[width=\textwidth]{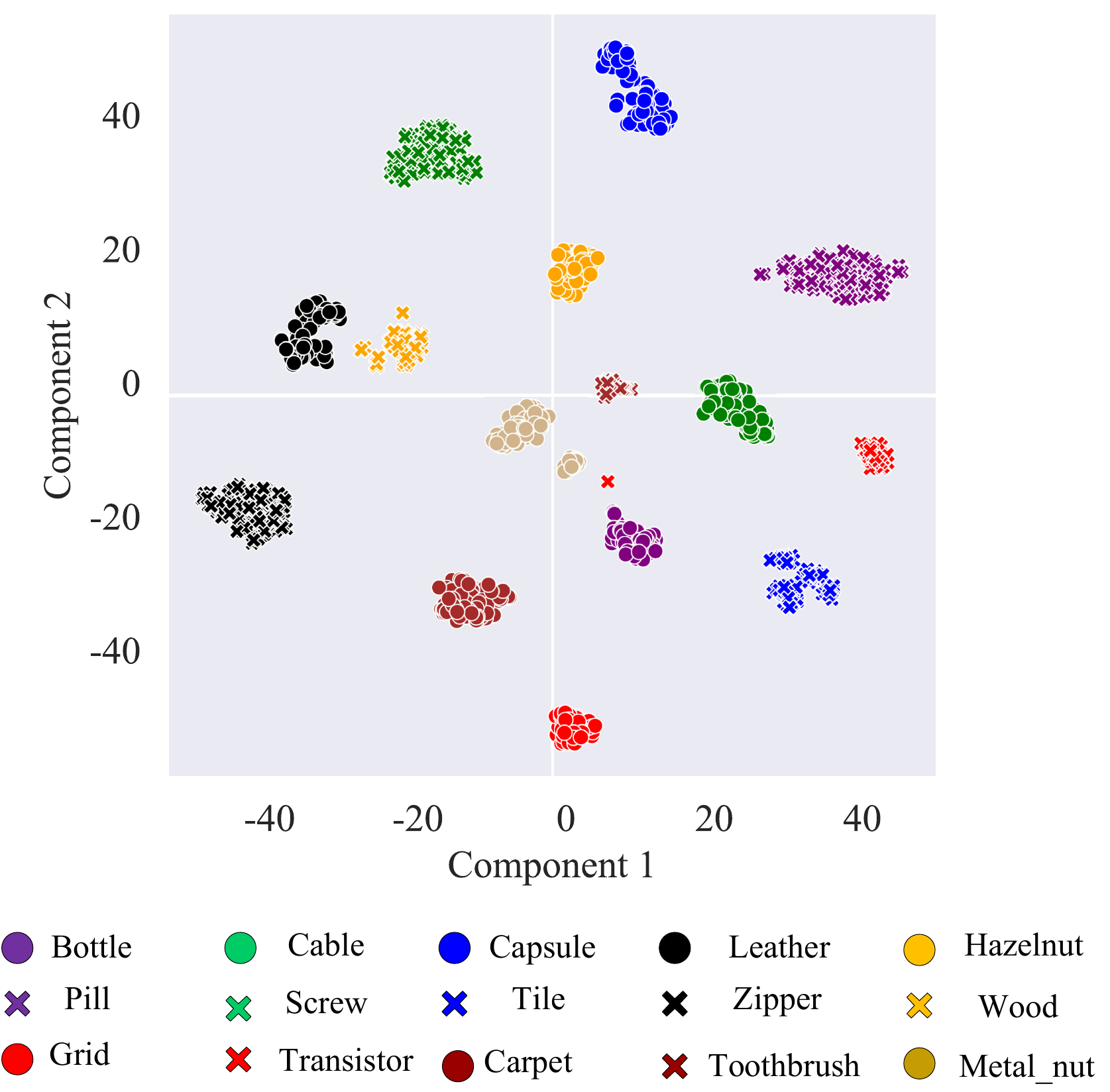}
		\caption{}
		\label{fig:sub1}
	\end{subfigure}
	\hfill
	\begin{subfigure}[b]{0.47\textwidth}
		\centering
		\includegraphics[width=\textwidth]{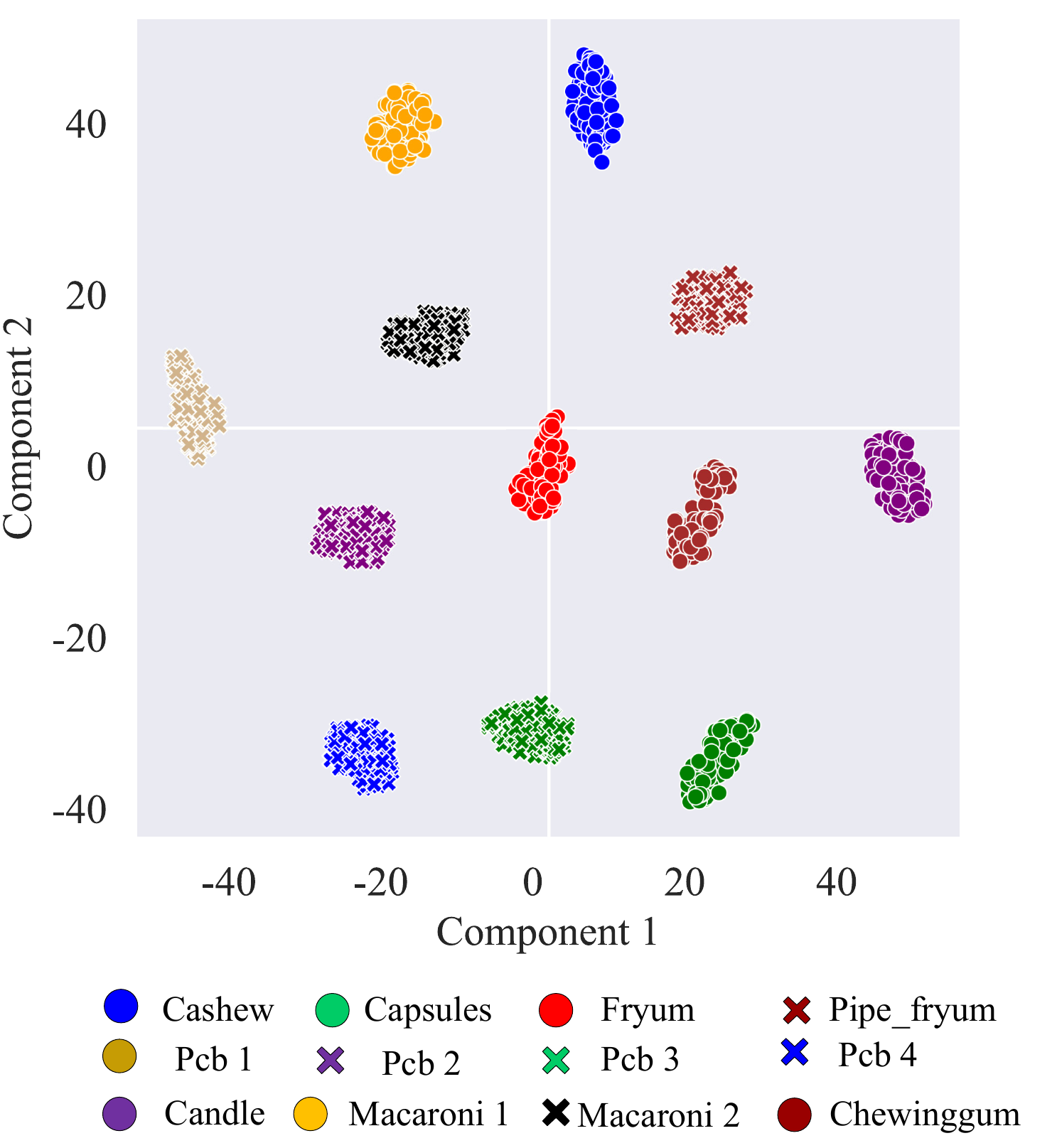}
		\caption{}
		\label{fig:sub2}
	\end{subfigure}
	\caption{Visualization of output text embeddings (classification weights) using t-SNE dimensionality reduction technique. (a) MVTec-AD dataset. (b) VisA dataset.}
	\label{fig10}
\end{figure}

\begin{table}[t]
	\caption{The experimental results when using different text prompt templates during testing. \textit{Prompt} 1 is the default template used during training, while \textit{Prompt} 2 to 5 are newly designed text prompt templates for testing. The best results are shown in \textbf{bold} and the second best results are \underline{underlined}.}
	\label{tab9}
	\centering
	\renewcommand{\arraystretch}{1.2}
	\resizebox{0.9\columnwidth}{!}
	{
		\begin{tabular}{>{\centering\arraybackslash}p{2cm}>{\centering\arraybackslash}p{7.6cm}*{6}{>{\centering\arraybackslash}p{1.1cm}}}
			\toprule
			&                                                              & \multicolumn{3}{c}{MVTec-AD}                                          & \multicolumn{3}{c}{VisA}                                              \\   \cmidrule(lr){3-8}
			&                                                              & AUROC                 & PRO                   & AP                    & AUROC                 & PRO                   & AP                    \\   \midrule
			\multirow{2}{*}{\textit{Prompt} 1} & a photo of a good $[z(x,v)]$                               & \multirow{2}{*}{92.0} & \multirow{2}{*}{\textbf{87.3}} & \multirow{2}{*}{\textbf{49.4}} & \multirow{2}{*}{95.7} & \multirow{2}{*}{90.0}   & \multirow{2}{*}{30.1} \\
			& a photo of a damaged $[z(x,v)]$                            &                       &                       &                       &                       &                       &                       \\  \midrule
			\multirow{2}{*}{\textit{Prompt} 2} & This is a good photo of $[z(x,v)]$                         & \multirow{2}{*}{\underline{92.2}} & \multirow{2}{*}{86.7} & \multirow{2}{*}{\underline{48.9}} & \multirow{2}{*}{95.9} & \multirow{2}{*}{\underline{90.3}} & \multirow{2}{*}{30.3} \\
			& This is a damaged photo of   $[z(x,v)]$                    &                       &                       &                       &                       &                       &                       \\  \midrule
			\multirow{2}{*}{\textit{Prompt} 3} & It is a photo of a $[z(x,v)]$ without damage               & \multirow{2}{*}{\textbf{92.4}} & \multirow{2}{*}{86.5} & \multirow{2}{*}{48.8} & \multirow{2}{*}{\underline{96.0}} & \multirow{2}{*}{\underline{90.3}} & \multirow{2}{*}{\underline{31.3}} \\
			& It is a photo of a $[z(x,v)]$   with damage                &                       &                       &                       &                       &                       &                       \\  \midrule
			\multirow{2}{*}{\textit{Prompt} 4} & There is not a damaged $[z(x,v)]$ in the photo             & \multirow{2}{*}{\textbf{92.4}} & \multirow{2}{*}{86.3} & \multirow{2}{*}{\underline{48.9}} & \multirow{2}{*}{\underline{96.0}} & \multirow{2}{*}{89.4} & \multirow{2}{*}{\textbf{31.7}} \\
			& There is a damaged $[z(x,v)]$ in   the photo               &                       &                       &                       &                       &                       &                       \\  \midrule
			\multirow{2}{*}{\textit{Prompt} 5} & It is a good, perfect and pristine picture of $[z(x,v)]$   & \multirow{2}{*}{92.1} & \multirow{2}{*}{\underline{87.0}} & \multirow{2}{*}{48.7} & \multirow{2}{*}{\textbf{96.1}} & \multirow{2}{*}{\textbf{90.5}} & \multirow{2}{*}{30.5} \\
			& It is a damaged, flawed and   broken picture of $[z(x,v)]$ &                       &                       &                       &                       &                       &          \\  \bottomrule           
		\end{tabular}
	}
\end{table}

In this subsection, we first compare the efficiency of different state-of-the-art methods. Subsequently, we visualize the output text embeddings to analyze the impact of visual context prompting.
\par 
\textbf{Analysis of model efficiency}.
In addition to anomaly segmentation performance, the efficiency of the model is also a focal point of our attention. Table \ref{tab8} reports the average inference time and the GPU cost per single image during the test stage. Despite not requiring auxiliary datasets for training, WinCLIP and AnVoL exhibit lower inference speed and segmentation performance compared to our VCP-CLIP. The inference time and GPU consumption of AnomalyGPT are substantial, far exceeding those of other methods, making it difficult to apply in real-world industrial scenarios. Compared to training-required methods like CoCoOp and APRIL-GAN, although the inference time of our VCP-CLIP slightly increases, it brings more gains in ZSAS performance.
\par
\textbf{Do text embeddings effectively integrate visual contexts?} Assuming the model's output normal and abnormal text embeddings are denoted as $g_n \in \mathbb{R}^{1\times C}$ and $g_a \in \mathbb{R}^{1\times C}$ respectively, and an image embedding of a patch as $x_p \in \mathbb{R}^{1\times C}$. Subsequently, if
$x_p$ is classified as abnormal, then we have: 
\begin{equation*}
	\widetilde{g}_n\widetilde{x}_p^T < \widetilde{g}_a\widetilde{x}_p^T  \Rightarrow  (\widetilde{g}_n - \widetilde{g}_a)\widetilde{x}_p^T < 0
\end{equation*}
where $\widetilde{(\cdot)}$ represents $L_2$-normalized operation along the embedding dimension. Then $g = \widetilde{g}_n - \widetilde{g}_a$ can be considered as classification weights (hyperplanes) for patch features. 
\par 
As shown in Fig. \ref{fig10}, we visualize the classification weights $g$ using the output text embedding (i.e., $ O_t^4$ in Equation 8) derived from the last Post-VCP module. Specifically, we first extract the output text embeddings for each input image after visual context prompting. Then, we apply t-SNE dimensionality reduction technique to the classification weights $g$ obtained from the text embeddings. Our observation reveals that the classification weights corresponding to images of the same product, across both the MVTec-AD and VisA datasets, are clustered together. This signifies that the output text embeddings fully integrate visual contexts, enabling the model to focus on the shared characteristics (domain information) inherent in the products. Additionally, we note that the classification weights for different images are distinctly separate, even if they belong to the same product type. This suggests that our VCP-CLIP concurrently learns the distinctions between images, allowing the text embeddings to dynamically adjust based on the input images. Therefore, compared to using unified text prompts, our VCP-CLIP showcases enhanced generalization capabilities for novel products by leveraging visual context prompting.
\par 
\textbf{Can VCP-CLIP generalize to other text prompt templates during testing?} Despite being trained on a fixed prompt template (\textit{Prompt} 1 in Table 3), VCP-CLIP still demonstrates strong ZSAS capabilities on other text prompt templates during testing. As shown in Table \ref{tab9}, we employ four different text prompts during testing, namely \textit{Prompt} $2\sim5$, which are distinct from those used during training.  To our surprise, when using new prompt templates such as \textit{Prompt} 3 and\textit{ Prompt} 5, some metrics have actually improved compared to using \textit{Prompt} 1. This can be attributed to the design of visual context prompting, which dynamically update text embeddings based on input images. 

\subsection{Results of zero-shot anomaly classification}
	\begin{table}[h]
		\centering
\caption{Performance comparison of zero-shot anomaly classification with image-level AUROC (\%) and AP (\%).}
\label{tableB}
\renewcommand{\arraystretch}{1.06}
\resizebox{0.5\columnwidth}{!}
{
	\begin{tabular}{>{\centering\arraybackslash}p{1.9cm}*{4}{>{\centering\arraybackslash}p{1.5cm}}}
		\toprule
		\multirow{2}{*}{Methods} & \multicolumn{2}{c}{MVTec-AD} & \multicolumn{2}{c}{VisA} \\  \cmidrule(lr){2-3}  \cmidrule(lr){4-5}
		& AUROC       & AP       & AUROC     & AP     \\  \midrule
		WinCLIP                  & 91.8           & 96.5        & 78.1         & 81.2      \\
		AnomalyGPT               & 75.4           & 86.1        & 58.0         & 64.8      \\
		APRIL-GAN                & 86.1           & 93.5        & 78.0         & 81.4      \\  
		\rowcolor{lightergray} % 设置橙色背景
		VCP-CLIP          & \textbf{92.1}           & \textbf{96.9}        & \textbf{83.8}         & \textbf{87.6}    \\  \bottomrule    
	\end{tabular}
}
\end{table}
In Table \ref{tableB}, we compare the classification performance of VCP-CLIP with other SOTA methods, including WinCLIP \cite{WinCLIP}, AnomalyGPT \cite{AnomalyGPT}, and APRIL-GAN \cite{VAND}. We present the results of taking the maximum value of the pixel-level score map as the anomaly classification score.
The results indicate that our zero-shot anomaly classification performance still surpasses other SOTA methods.

\section{Datasets}  \label{secC}
In this section, we provide a detailed introduction to ten real industrial datasets used in this paper. More details are as follows:
\begin{itemize}[label=$\diamond$]
	\item \textbf{MVTec-AD} \cite{MVTec}. It contains 5354 color images with resolutions ranging from 700 to 1024, which are used for industrial anomaly detection. It consists of 15 types of products containing object and texture categories with pixel-level annotations.
	\item \textbf{VisA} \cite{VisA}. It contains 10,821 color images with resolutions ranging from 960 to 1500, used for industrial anomaly detection. It comprises 12 types of products containing object category with pixel-level annotations. 
	\item \textbf{BSD} \cite{BSD}. It is a ball screw drive nut database that consists of 1104  color images with a resolution of 2592 × 1944, showing areas with and without pitting(s). The dataset contains 710 images without pitting and 394 images with pitting. In our study, we only use the defective images for evaluation.
	\item \textbf{GC} \cite{GC}. It is a surface defect dataset collected in a real industry. It contains ten types of steel surface defects and includes 3570 gray-scale images. The original annotations are in the form of bounding boxes. In the experiments, the MobileSAM \cite{mobileSAM} model is used for assisted annotation to obtain segmentation ground truth.
	\item  \textbf{KSDD2} \cite{KSDD2}. It is a dataset that captured in a controlled environment with 230 pixels wide and 630 pixels high. The dataset has 2085 negative and 246 positive samples in the train, and 894 negative and 110 positive samples in the test subset. Defects are annotated with fine-grained segmentation masks and vary in shape, size and color, rangin gfrom small scratches and minor spots to large surface imperfections. 
	\item \textbf{MSD} \cite{MSD}. It is a mobile phone screen surface defect dataset that consists of 1200 images. The defects are made by ourselves and pixel-level labeled by labelme. The images are collected by an industrial camera and the resolution is 1920 × 1080.
	\item \textbf{Road} \cite{Road}. It is a road crack dataset that is composed of 118 images, which can generally reflect urban road surface condition in Beijing, China. Each image has hand labeled ground truth contours. The width of the images ranges from 1 to 3 mm. Tese images contain noises such as shadows, oil spots and water stains. 
	\item \textbf{RSDD} \cite{RSDD}. This dataset contains  two style data sets from the China Academy of Railway Sciences, collected using a linear array camera: 1) Type-I RSDD data set contains 67 images captured from express rails; 2) Type-II RSDD data set contains 128 images captured from common/heavy haul rails.Note that each image from both data sets contains at least one defect.
	\item \textbf{BTech} \cite{BTech}. It is an anomaly detection dataset. It consists of three subdatasets. Among them, Product 1 has a resolution of 1600 × 1600 pixels, Product 2 has a resolution of 600 × 600 pixels, and Product 3 has a resolution of 800 × 600 pixels. Product 1, 2, and 3 have 400, 1000, and 399 training images, respectively. 
	\item \textbf{DAGM} \cite{DAGM}. It is a manually generated texture dataset, consisting of a total of 10 classes, with image size of 512×512 pixels. The original annotations were weakly supervised annotations, where the defect areas are in the form of ellipses. Therefore, in the experiments, MobileSAM \cite{mobileSAM} is used for fine annotation.
\end{itemize}
\par 
In this experiment, the image size used for the experimental evaluation in \cite{BSD, GC, KSDD2, MSD, Road, RSDD, BTech, DAGM} was uniformly 512 x 512 pixels, all of which were randomly selected and cropped from the original dataset. Subsequently, all eight processed industrial datasets mentioned above will be open source.
\newpage
\section{Detailed ZSAS results}   \label{secD}
In this section, we provide detailed quantitative and qualitative results specific to the products in our VCP-CLIP.
\subsection{Detailed quantitative results in different products}
\begin{table}[H]
	\caption{Comparison of different products in terms of AUROC(\%) on MVTec-AD. The best results are shown in \textbf{bold}.}
	\label{tab10}
	\centering
	\renewcommand{\arraystretch}{1.1}
	\resizebox{0.85\columnwidth}{!}
	{
		\begin{tabular}{>{\centering\arraybackslash}p{1.8cm} *{3}{>{\centering\arraybackslash}p{1.4cm}}*{2}{>{\centering\arraybackslash}p{2cm}} *{2}{>{\centering\arraybackslash}p{1.6cm}}}
			\toprule
			Product & WinCLIP & AnVoL & CoCoOp & AnomalyGPT & APRIL-GAN & Baseline & VCP-CLIP \\   \midrule
			Bottle             & 89.5    & --    & 87.9   & 89.1       & 83.5      & 91.7     & \textbf{94.1}     \\
			Cable              & 77.0    & --    & 67.6   & 52.2       & 72.2      & \textbf{79.4}     & 78.6     \\
			Capsule            & 86.9    & --    & 92.3   & 88.6       & 92.0      & 95.8     & \textbf{96.6}     \\
			Carpet             & 95.4    & --    & 97.2   & 97.0       & 98.4      & 99.3     & \textbf{99.6}     \\
			Grid               & 82.2    & --    & 96.4   & 68.0       & 95.8      & \textbf{98.2}     & 97.9     \\
			Hazelnut           & 94.3    & --    & 96.5   & 92.0       & 96.1      & 97.4     & \textbf{97.7}     \\
			Leather            & 96.7    & --    & \textbf{99.6}   & 73.2       & 99.1      & \textbf{99.6}     & \textbf{99.6}     \\
			Metal\_nut         & 61.0    & --    & \textbf{78.8}   & 65.7       & 65.5      & 75.4     & 74.0     \\
			Pill               & 80.0    & --    & 73.4   & \textbf{93.7}       & 76.2      & 57.5     & 90.1     \\
			Screw              & 89.6    & --    & 97.5   & 96.0       & 97.8      & 98.1     & \textbf{98.5}     \\
			Tile               & 77.6    & --    & 88.5   & 46.6       & 92.7      & 92.9     & \textbf{93.5}     \\
			Toothbrush         & 86.9    & --    & 94.7   & 92.6       & \textbf{95.8}      & 93.2     & 94.7     \\
			Transistor         & \textbf{74.7}    & --    & 63.4   & 61.4       & 62.4      & 67.6     & 69.8     \\
			Wood               & 93.4    & --    & 95.2   & 88.7       & 95.8      & 96.7     & \textbf{97.6}     \\
			Zipper             & 91.6    & --    & 93.9   & 87.6       & 91.1      & 94.8     & \textbf{98.1}     \\   \midrule
			Mean               & 85.1    & 90.6  & 88.2   & 79.5       & 87.6      & 89.2     & \textbf{92.0}    \\ \bottomrule
		\end{tabular}
	}
\end{table}
\begin{table}[!h]
	\caption{Comparison of different products in terms of PRO(\%) on MVTec-AD. The best results are shown in \textbf{bold}.}
	\label{tab11}
	\centering
	\renewcommand{\arraystretch}{1.1}
	\resizebox{0.85\columnwidth}{!}
	{
		\begin{tabular}{>{\centering\arraybackslash}p{1.8cm} *{3}{>{\centering\arraybackslash}p{1.4cm}}*{2}{>{\centering\arraybackslash}p{2cm}} *{2}{>{\centering\arraybackslash}p{1.6cm}}}
			\toprule
			Product & WinCLIP & AnVoL & CoCoOp & AnomalyGPT & APRIL-GAN & Baseline & VCP-CLIP \\   \midrule
			Bottle           & 76.4    & --    & 83.0   & 65.4       & 45.6      & 84.6     & \textbf{88.5}     \\ 
			Cable            & 42.9    & --    & 55.9   & 41.7       & 25.7      & 63.7     & \textbf{67.7}     \\
			Capsule          & 62.1    & --    & 89.0   & 48.2       & 51.3      & 91.6     & \textbf{94.0}     \\
			Carpet           & 84.1    & --    & 93.2   & 22.8       & 48.5      & 97.7     & \textbf{98.4}     \\
			Grid             & 57.0    & --    & 87.4   & 20.7       & 31.6      & \textbf{94.6}     & 92.4     \\
			Hazelnut         & 81.6    & --    & 82.3   & 51.7       & 70.3      & \textbf{91.5}     & 84.9     \\
			Leather          & 91.1    & --    & 99.2   & 30.4       & 72.4      & \textbf{99.3}     & 99.2     \\
			Metal\_nut       & 31.8    & --    & \textbf{78.6}   & 57.8       & 38.4      & 73.0     & 73.4     \\
			Pill             & 65.0    & --    & 92.5   & 69.7       & 65.4      & 91.3     & \textbf{94.9}     \\
			Screw            & 68.5    & --    & 91.2   & 74.8       & 67.1      & 92.4     & \textbf{93.2}     \\
			Tile             & 51.2    & --    & 81.0   & 10.5       & 26.7      & 89.9     & \textbf{90.6}     \\
			Toothbrush       & 67.7    & --    & 83.8   & 75.9       & 54.5      & 87.4     & \textbf{87.8}     \\
			Transistor       & 43.4    & --    & 54.6   & 43.9       & 21.3      & 54.7     & \textbf{56.1}     \\
			Wood             & 74.1    & --    & 93.4   & 35.7       & 31.1      & 95.5     & \textbf{95.7}     \\
			Zipper           & 71.7    & --    & 82.2   & 39.4       & 10.7      & 80.6     & \textbf{92.2}     \\ \midrule
			Mean             & 64.6    & 77.8  & 83.2   & 45.9       & 44.0      & 85.8     & \textbf{87.3}     \\ \bottomrule
		\end{tabular}
	}
\end{table}

\begin{table}[!h]
	\caption{Comparison of different products in terms of AP(\%) on MVTec-AD. The best results are shown in \textbf{bold}}
	\label{tab12}
	\centering
	\renewcommand{\arraystretch}{1.2}
	\resizebox{0.85\columnwidth}{!}
	{
		\begin{tabular}{>{\centering\arraybackslash}p{1.8cm} *{3}{>{\centering\arraybackslash}p{1.4cm}}*{2}{>{\centering\arraybackslash}p{2cm}} *{2}{>{\centering\arraybackslash}p{1.6cm}}}
			\toprule
			Product & WinCLIP & AnVoL & CoCoOp & AnomalyGPT & APRIL-GAN & Baseline & VCP-CLIP \\   \midrule
			Bottle          & 49.8    & 55.7  & 59.2   & 50.1       & 53.0      & 59.9     & \textbf{67.2}     \\
			Cable           & 6.2     & \textbf{20.4}  & 7.2    & 17.2       & 18.2      & 17.8     & 19.5     \\
			Capsule         & 8.6     & 16.0  & 27.1   & 12.7       & 29.6      & 26.1     & \textbf{35.6}     \\
			Carpet          & 25.9    & 61.4  & 68.9   & 61.2       & 67.5      & 80.2     & \textbf{84.3}     \\
			Grid            & 5.7     & 16.0  & 38.5   & 5.9        & 36.5      & \textbf{43.7}     & 42.8     \\
			Hazelnut        & 33.3    & 33.2  & 49.8   & 30.0       & 49.7      & 56.9     & \textbf{59.0}     \\
			Leather         & 20.4    & 34.3  & 62.4   & 3.6        & 52.3      & 64.0     & \textbf{65.0}     \\
			Metal\_nut      & 10.8    & 17.0  & \textbf{33.1}   & 20.1       & 25.9      & 26.9     & 25.0     \\
			Pill            & 7.0     & 12.8  & 23.5   & \textbf{39.6}       & 23.6      & 27.1     & 34.3     \\
			Screw           & 5.4     & 1.0   & 28.7   & 11.5       & 33.7      & 35.8     & \textbf{38.1}     \\
			Tile            & 21.2    & 48.4  & 60.6   & 9.9        & 66.3      & 71.8     & \textbf{76.2}     \\
			Toothbrush      & 5.5     & 20.6  & 19.6   & 30.8       & 43.2      & 30.9     & \textbf{35.6}     \\
			Transistor      & \textbf{20.2}    & 16.2  & 11.0   & 10.9       & 11.7      & 13.6      & 14.2     \\
			Wood            & 32.9    & 57.6  & 68.0   & 26.4       & 61.8      & 71.7     & \textbf{74.2}     \\
			Zipper          & 19.4    & 10.8  & 48.3   & 25.5       & 38.7      & 52.2     & \textbf{70.4}     \\  \midrule
			Mean            & 18.2    & 28.1  & 40.4   & 23.7       & 40.8      & 45.2     & \textbf{49.4}     \\ \bottomrule
		\end{tabular}
	}
\end{table}

\begin{table}[!h]
	\caption{Comparison of different products in terms of AUROC(\%) on VisA. The best results are shown in \textbf{bold}.}
	\label{tab13}
	\centering
	\renewcommand{\arraystretch}{1.2}
	\resizebox{0.85\columnwidth}{!}
	{
		\begin{tabular}{>{\centering\arraybackslash}p{1.8cm} *{3}{>{\centering\arraybackslash}p{1.4cm}}*{2}{>{\centering\arraybackslash}p{2cm}} *{2}{>{\centering\arraybackslash}p{1.6cm}}}
			\toprule
			Product & WinCLIP & AnVoL & CoCoOp & AnomalyGPT & APRIL-GAN & Baseline & VCP-CLIP \\   \midrule
			Candle         & 88.9    & --    & 98.6   & 91.3       & 97.8      & \textbf{99.2}     & \textbf{99.2}     \\
			Capsules       & 81.6    & --    & 96.1   & 88.8       & 97.5      & 97.8     & \textbf{98.7}     \\
			Cashew         & 84.7    & --    & \textbf{93.2}   & 89.8       & 86.0      & 91.2     & 93.1     \\
			Chewinggum     & 93.3    & --    & 99.4   & 96.0       & \textbf{99.5}      & 99.4     & \textbf{99.5}     \\
			Fryum          & 88.5    & --    & 93.3   & 89.7       & 92.0      & 94.4     & \textbf{94.6}     \\
			Macaroni1      & 70.9    & --    & 99.3   & 92.4       & 98.8      & 99.3     & \textbf{99.6}     \\
			Macaroni2      & 59.3    & --    & 98.6   & 90.6       & 97.8      & 97.9     & \textbf{98.7}     \\
			Pcb1           & 61.2    & --    & 92.6   & 88.4       & 92.7      & \textbf{92.8}     & 92.1     \\
			Pcb2           & 71.6    & --    & 90.4   & 82.6       & 89.8      & \textbf{92.5}     & 92.1     \\
			Pcb3           & 85.3    & --    & 89.3   & 86.7       & 88.4      & \textbf{90.1}     & 89.2     \\
			Pcb4           & 94.4    & --    & 93.2   & 92.1       & 94.6      & \textbf{95.8}     & 95.6     \\
			Pipe\_fryum    & 75.4    & --    & 95.1   & 95.3       & 96.0      & 94.9     & \textbf{96.4}     \\   \midrule
			Mean           & 79.6    & 91.4  & 94.9   & 90.3       & 94.2      & 95.5     & \textbf{95.7}      \\ \bottomrule
		\end{tabular}
	}
\end{table}

\begin{table}[!h]
	\caption{Comparison of different products in terms of PRO(\%) on VisA. The best results are shown in \textbf{bold}.}
	\label{tab14}
	\centering
	\renewcommand{\arraystretch}{1.1}
	\resizebox{0.85\columnwidth}{!}
	{
		\begin{tabular}{>{\centering\arraybackslash}p{1.8cm} *{3}{>{\centering\arraybackslash}p{1.4cm}}*{2}{>{\centering\arraybackslash}p{2cm}} *{2}{>{\centering\arraybackslash}p{1.6cm}}}
			\toprule
			Product & WinCLIP & AnVoL & CoCoOp & AnomalyGPT & APRIL-GAN & Baseline & VCP-CLIP \\   \midrule
			Candle       & 83.5    & --    & 95.2   & 36.7       & 92.5      & \textbf{96.2}     & \textbf{96.2}     \\
			Capsules     & 35.3    & --    & 86.1   & 60.8       & 86.7      & 87.8     & \textbf{91.2}     \\
			Cashew       & 76.4    & --    & 93.6   & 63.6       & 91.7      & 93.1     & \textbf{95.6}     \\
			Chewinggum   & 70.4    & --    & 89.1   & 54.8       & 87.3      & \textbf{92.}2     & \textbf{92.2}     \\
			Fryum        & 77.4    & --    & 88.7   & 64.7       & 89.7      & 90.9     & \textbf{92.2}     \\
			Macaroni1    & 34.3    & --    & 95.9   & 71.3       & 93.2      & 96.0     & \textbf{97.5}     \\
			Macaroni2    & 21.4    & --    & 89.2   & 67.4       & 82.3      & 85.7     & \textbf{90.5}     \\
			Pcb1         & 26.3    & --    & 85.8   & 59.8       & 87.5      & \textbf{89.1}     & 88.1     \\
			Pcb2         & 37.2    & --    & 75.3   & 47.8       & 75.6      & 78.5     & \textbf{79.1}     \\
			Pcb3         & 56.1    & --    & 79.3   & 58.7       & 77.8      & \textbf{80.7}     & 78.9     \\
			Pcb4         & 80.4    & --    & 82.9   & 68.6       & 86.8      & 89.2     & \textbf{89.7}     \\
			Pipe\_fryum  & 82.3    & --    & 94.9   & 83.3       & 90.9      & 96.3     & \textbf{96.5}     \\  \midrule
			Mean         & 56.8    & 75.0  & 88.0   & 61.5       & 86.8      & 89.6     & \textbf{90.7}    \\ \bottomrule
		\end{tabular}
	}
\end{table}

\begin{table}[!h]
	\caption{Comparison of different products in terms of AP(\%) on VisA. The best results are shown in \textbf{bold}.}
	\label{tab15}
	\centering
	\renewcommand{\arraystretch}{1.1}
	\resizebox{0.85\columnwidth}{!}
	{
		\begin{tabular}{>{\centering\arraybackslash}p{1.8cm} *{3}{>{\centering\arraybackslash}p{1.4cm}}*{2}{>{\centering\arraybackslash}p{2cm}} *{2}{>{\centering\arraybackslash}p{1.6cm}}}
			\toprule
			Product & WinCLIP & AnVoL & CoCoOp & AnomalyGPT & APRIL-GAN & Baseline & VCP-CLIP \\   \midrule
			Candle      & 2.4     & 9.1   & 32.7   & 15.1       & 29.9      & 40.7     & \textbf{41.6}     \\
			Capsules    & 1.4     & 1.0   & 34.0   & 19.6       & 40.0      & 42.1     & \textbf{51.5}     \\
			Cashew      & 4.8     & 9.1   & 22.1   & 5.6        & 15.1      & 20.0     & \textbf{26.5}     \\
			Chewinggum  & 24.0    & 50.2  & 82.4   & 46.1       & 83.6      & 79.9     & \textbf{85.3}     \\
			Fryum       & 11.1    & \textbf{31.2}  & 23.3   & 17.8       & 22.1      & 26.8     & 28.2     \\
			Macaroni1   & 0.03    & 0.2   & 25.2   & 3.1        & 24.8      & 20.5     & \textbf{25.7}     \\
			Macaroni2   & 0.02    & 0.1   & \textbf{8.8}    & 0.4        & 6.8       & 1.9      & 3.3      \\
			Pcb1        & 0.4     & \textbf{24.7}  & 5.4    & 3.5        & 8.4       & 7.9      & 8.6      \\
			Pcb2        & 0.4     & 1.4   & 9.1    & 1.9        & 15.4      & 15.5     & \textbf{15.9}     \\
			Pcb3        & 0.7     & 1.8   & 16.0   & 3.5        & 14.1      & \textbf{20.6}     & 15.6     \\
			Pcb4        & 15.5    & 19.0  & 20.1   & 16.7       & 24.9      & \textbf{33.2}     & 31.9     \\
			Pipe\_fryum & 4.4     & 4.2   & 19.2   & 25.8       & 23.6      & 18.6     & \textbf{26.3}     \\  \midrule
			Mean        & 5.4     & 12.7  & 24.8   & 13.3       & 25.7      & 27.3     & \textbf{30.1}      \\ \bottomrule
		\end{tabular}
	}
\end{table}
\clearpage
\subsection{Detailed qualitative results in different products}
\begin{figure}[!h]
	\centering
	\includegraphics[width=1\columnwidth]{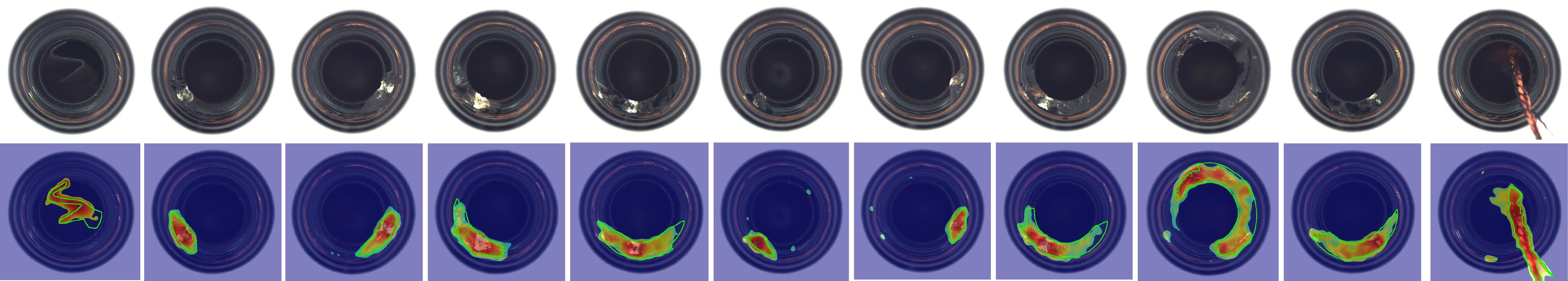}
	\caption{Segmentation results for the product, bottle, on MVTec AD. The first row depicts the original image, while the second row shows the anomaly segmentation results, with the regions encircled in green representing the ground truth.}
	\label{fig11}
\end{figure} 
\begin{figure}[!h]
	\centering
	\includegraphics[width=1\columnwidth]{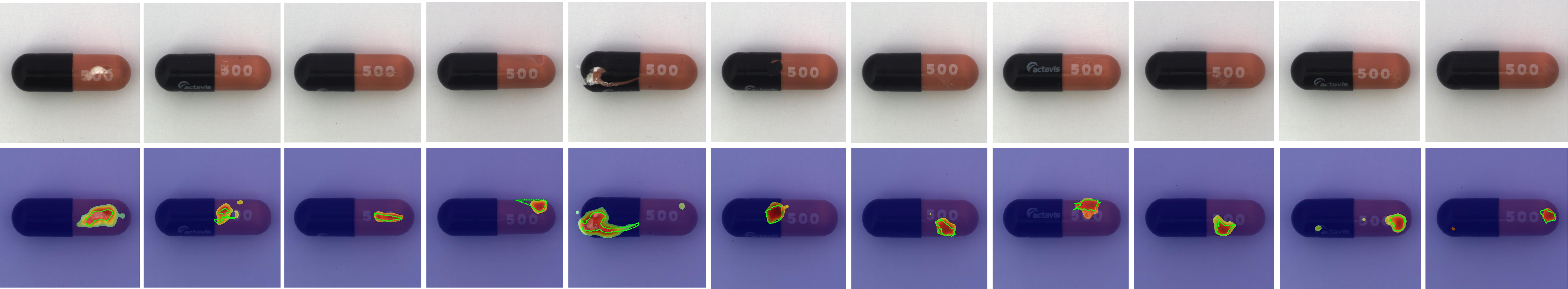}
	\caption{Segmentation results for the product, capsule, on MVTec AD. The first row depicts the original image, while the second row shows the anomaly segmentation results, with the regions encircled in green representing the ground truth.}
	\label{fig12}
\end{figure} 
\begin{figure}[!h]
	\centering
	\includegraphics[width=1\columnwidth]{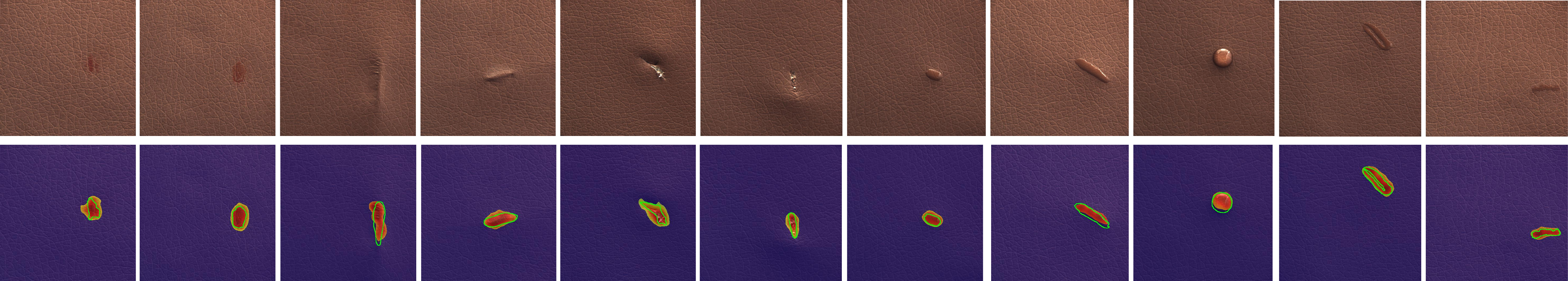}
	\caption{Segmentation results for the product, leather, on MVTec AD. The first row depicts the original image, while the second row shows the anomaly segmentation results, with the regions encircled in green representing the ground truth.}
	\label{fig13}
\end{figure} 
\begin{figure}[!h]
	\centering
	\includegraphics[width=1\columnwidth]{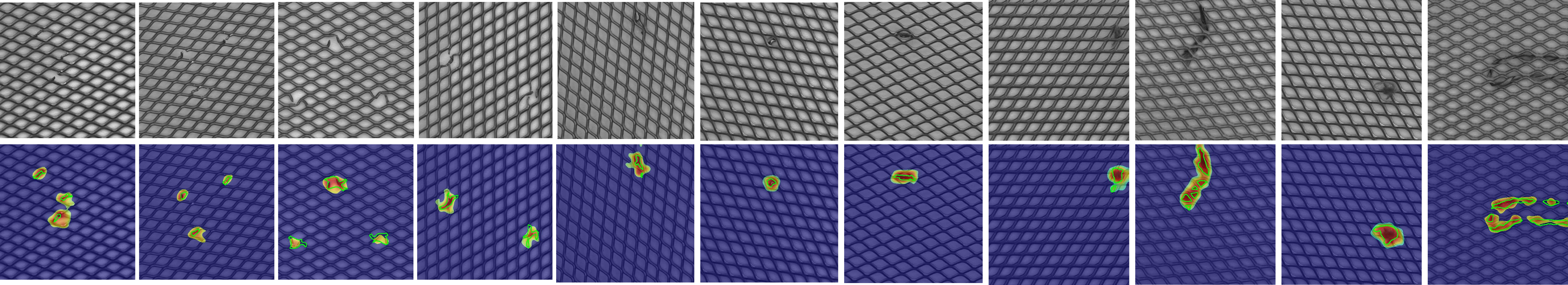}
	\caption{Segmentation results for the product, grid, on MVTec AD. The first row depicts the original image, while the second row shows the anomaly segmentation results, with the regions encircled in green representing the ground truth.}
	\label{fig14}
\end{figure} 
\begin{figure}[!h]
	\centering
	\includegraphics[width=1\columnwidth]{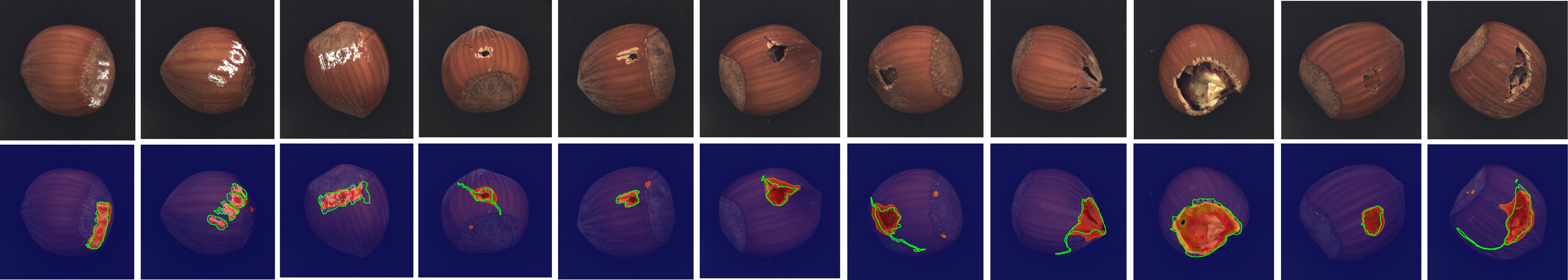}
	\caption{Segmentation results for the product, hazelnut, on MVTec AD. The first row depicts the original image, while the second row shows the anomaly segmentation results, with the regions encircled in green representing the ground truth.}
	\label{fig15}
\end{figure} 
\begin{figure}[!h]
	\centering
	\includegraphics[width=1\columnwidth]{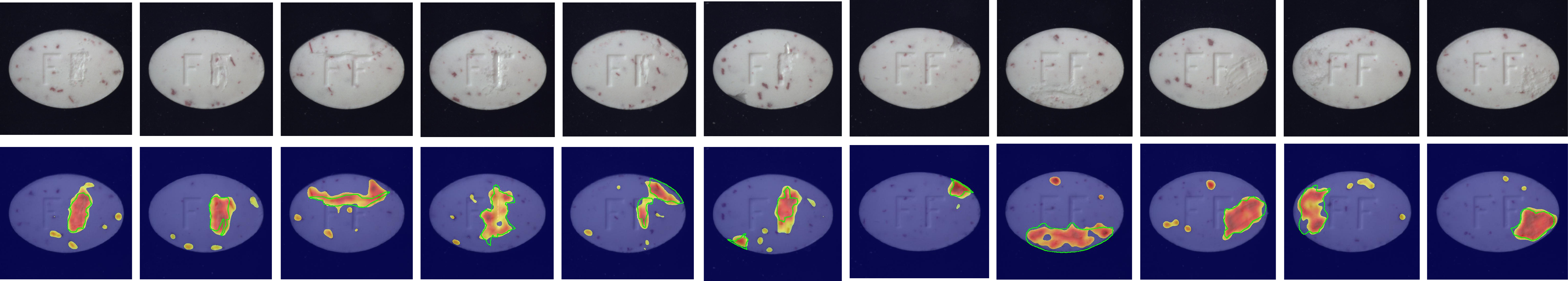}
	\caption{Segmentation results for the product, pill, on MVTec AD. The first row depicts the original image, while the second row shows the anomaly segmentation results, with the regions encircled in green representing the ground truth.}
	\label{fig16}
\end{figure} 
\begin{figure}[!h]
	\centering
	\includegraphics[width=1\columnwidth]{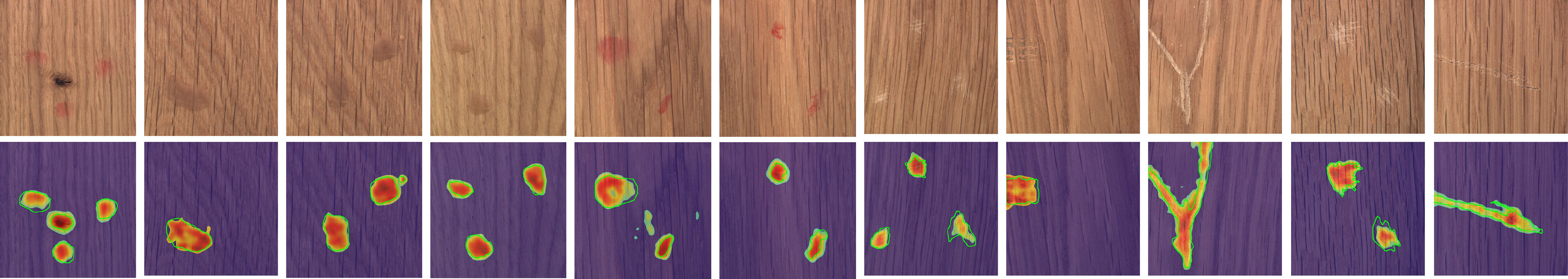}
	\caption{Segmentation results for the product, wood, on MVTec AD. The first row depicts the original image, while the second row shows the anomaly segmentation results, with the regions encircled in green representing the ground truth.}
	\label{fig17}
\end{figure} 
\begin{figure}[!h]
	\centering
	\includegraphics[width=1\columnwidth]{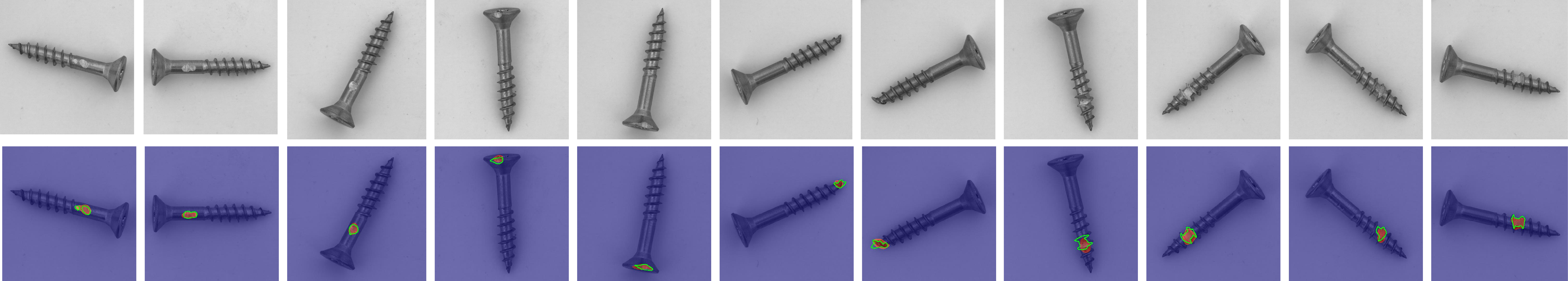}
	\caption{Segmentation results for the product, screw, on MVTec AD. The first row depicts the original image, while the second row shows the anomaly segmentation results, with the regions encircled in green representing the ground truth.}
	\label{fig18}
\end{figure} 
\begin{figure}[!h]
	\centering
	\includegraphics[width=1\columnwidth]{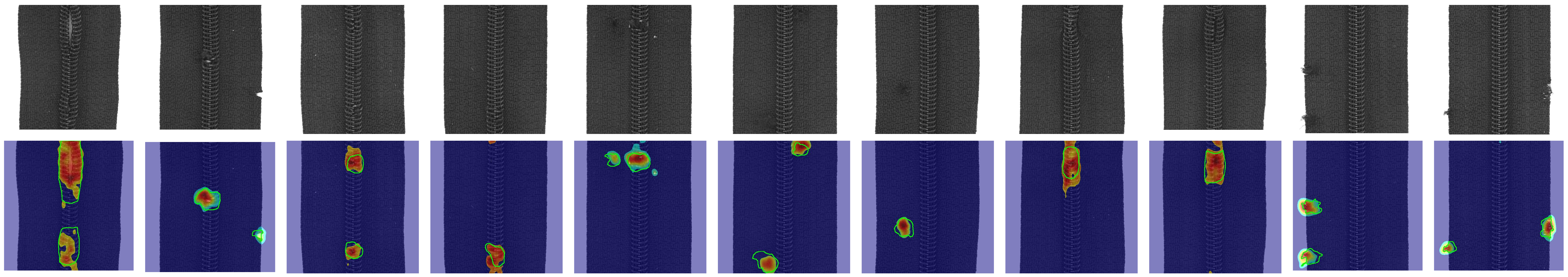}
	\caption{Segmentation results for the product, zipper, on MVTec AD. The first row depicts the original image, while the second row shows the anomaly segmentation results, with the regions encircled in green representing the ground truth.}
	\label{fig19}
\end{figure} 
\begin{figure}[h]
	\centering
	\includegraphics[width=1\columnwidth]{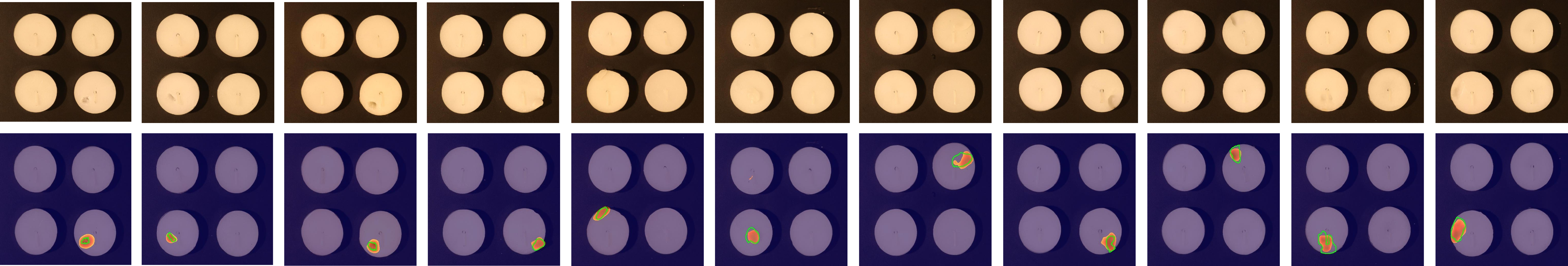}
	\caption{Segmentation results for the product, candle, on VisA. The first row depicts the original image, while the second row shows the anomaly segmentation results, with the regions encircled in green representing the ground truth.}
	\label{fig20}
\end{figure} 
\begin{figure}[!h]
	\centering
	\includegraphics[width=1\columnwidth]{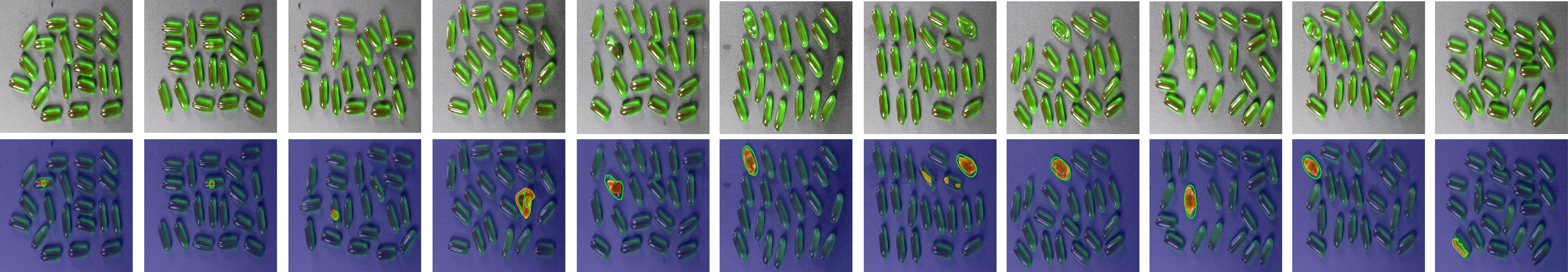}
	\caption{Segmentation results for the product, capsules, on VisA. The first row depicts the original image, while the second row shows the anomaly segmentation results, with the regions encircled in green representing the ground truth.}
	\label{fig21}
\end{figure} 

\begin{figure}[!h]
	\centering
	\includegraphics[width=1\columnwidth]{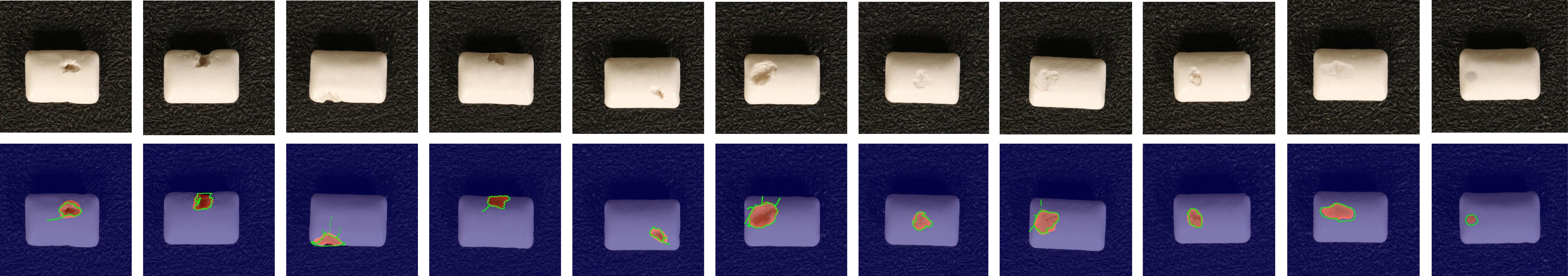}
	\caption{Segmentation results for the product, chewinggum, on VisA. The first row depicts the original image, while the second row shows the anomaly segmentation results, with the regions encircled in green representing the ground truth.}
	\label{fig22}
\end{figure} 

\begin{figure}[h]
	\centering
	\includegraphics[width=1\columnwidth]{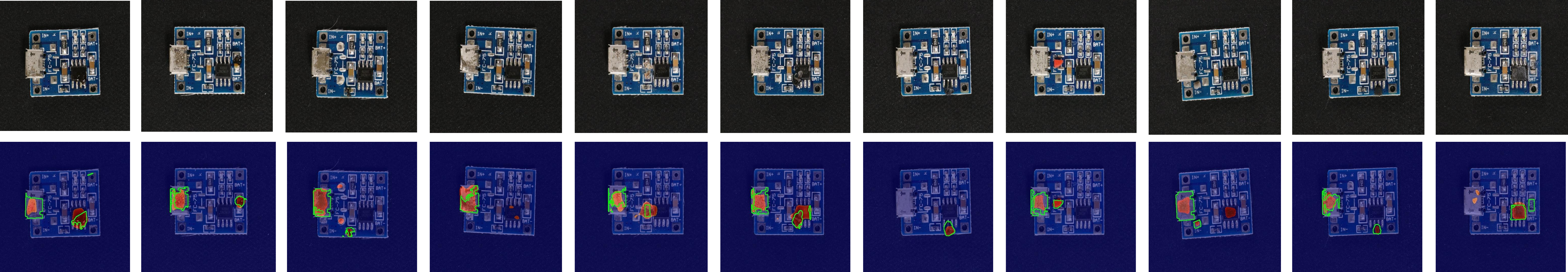}
	\caption{Segmentation results for the product, pcb4, on VisA. The first row depicts the original image, while the second row shows the anomaly segmentation results, with the regions encircled in green representing the ground truth.}
	\label{fig23}
\end{figure} 
\begin{figure}[!h]
	\centering
	\includegraphics[width=1\columnwidth]{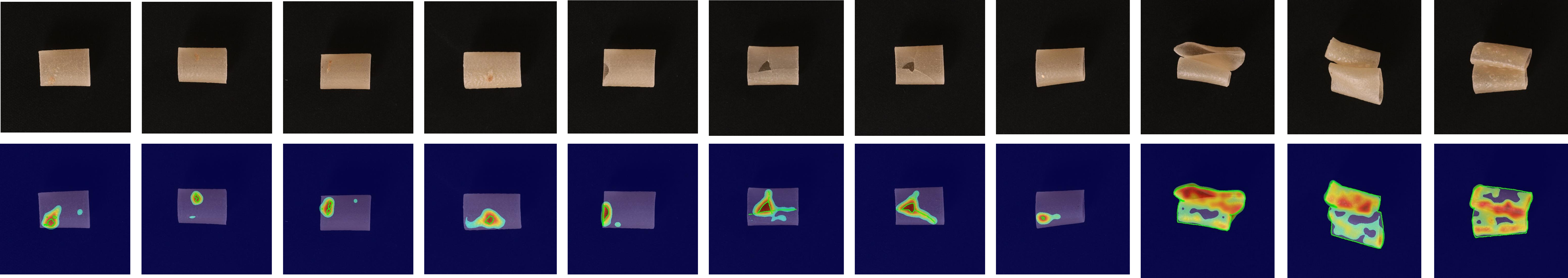}
	\caption{Segmentation results for the product, pipe\_fryum, on VisA. The first row depicts the original image, while the second row shows the anomaly segmentation results, with the regions encircled in green representing the ground truth.}
	\label{fig24}
\end{figure}
\begin{figure}[!h]
	\centering
	\includegraphics[width=1\columnwidth]{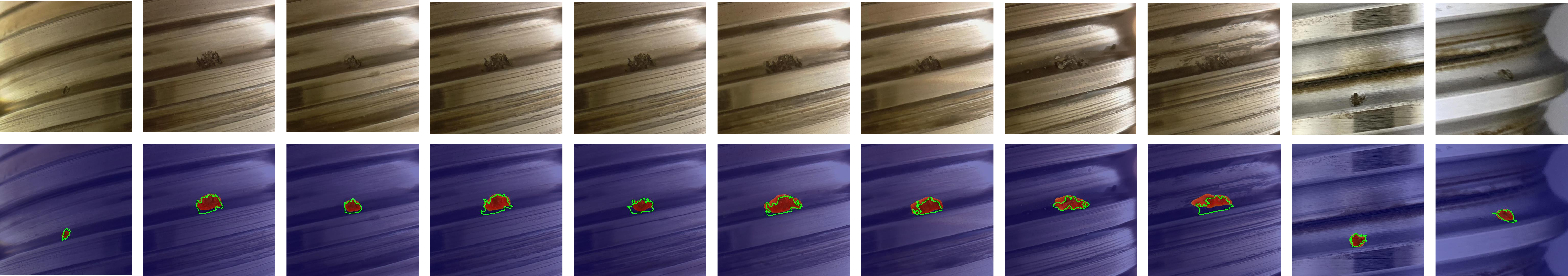}
	\caption{Segmentation results on the BSD datasets. The first row depicts the original image, while the second row shows the anomaly segmentation results, with the regions encircled in green representing the ground truth.}
	\label{fig25}
\end{figure}
\begin{figure}[h]
	\centering
	\includegraphics[width=1\columnwidth]{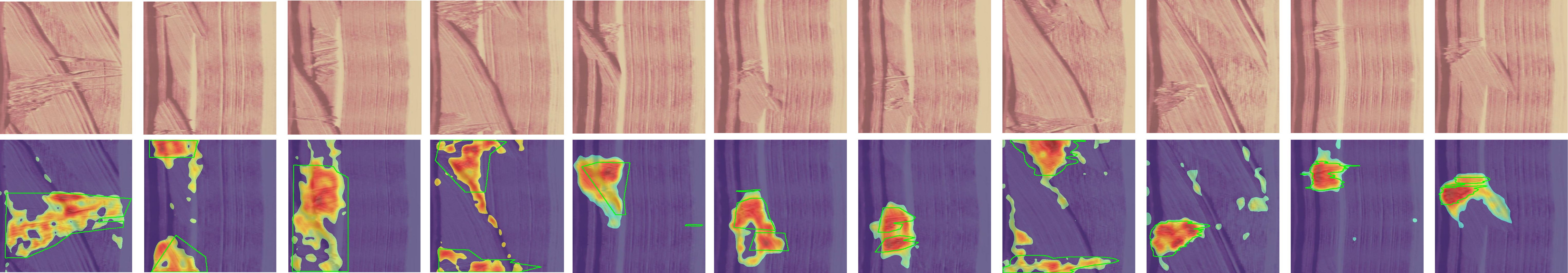}
	\caption{Segmentation results for the product, wood, on BTech. The first row depicts the original image, while the second row shows the anomaly segmentation results, with the regions encircled in green representing the ground truth.}
	\label{fig26}
\end{figure}
\begin{figure}[!h]
	\centering
	\includegraphics[width=1\columnwidth]{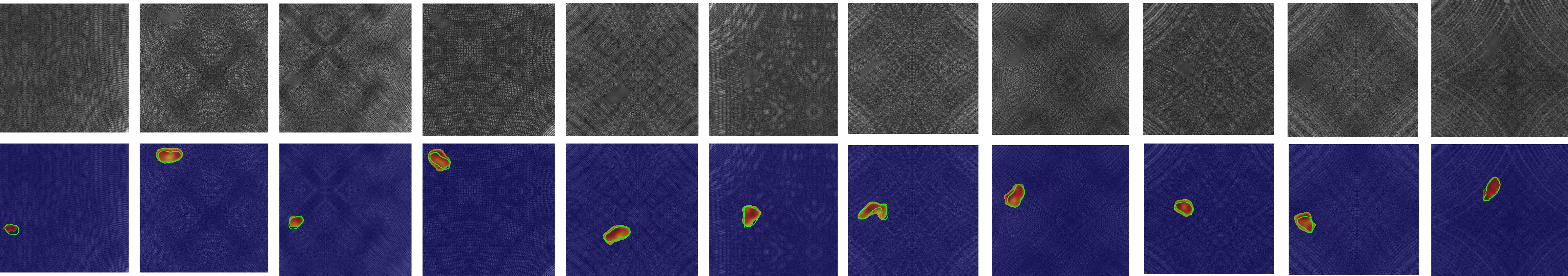}
	\caption{Segmentation results for the product, Class 1, on DAGM. The first row depicts the original image, while the second row shows the anomaly segmentation results, with the regions encircled in green representing the ground truth.}
	\label{fig27}
\end{figure}
\begin{figure}[!h]
	\centering
	\includegraphics[width=1\columnwidth]{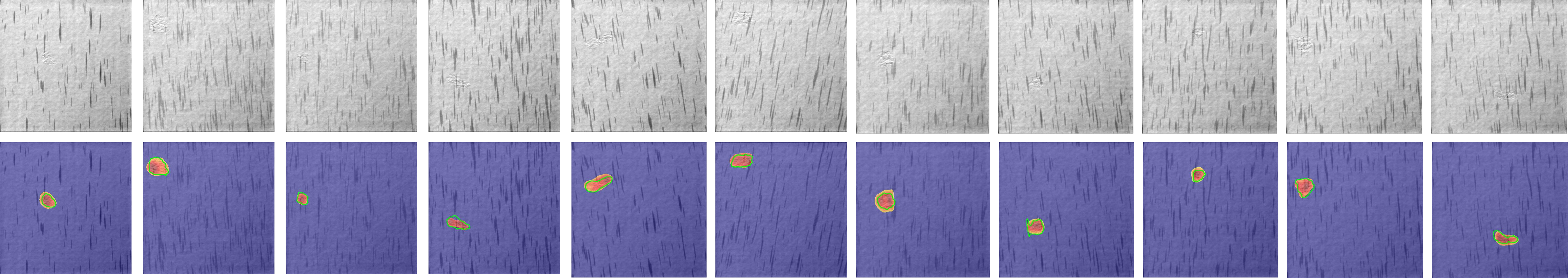}
	\caption{Segmentation results for the product, Class 7, on DAGM. The first row depicts the original image, while the second row shows the anomaly segmentation results, with the regions encircled in green representing the ground truth.}
	\label{fig28}
\end{figure}
\begin{figure}[!h]
	\centering
	\includegraphics[width=0.96\columnwidth]{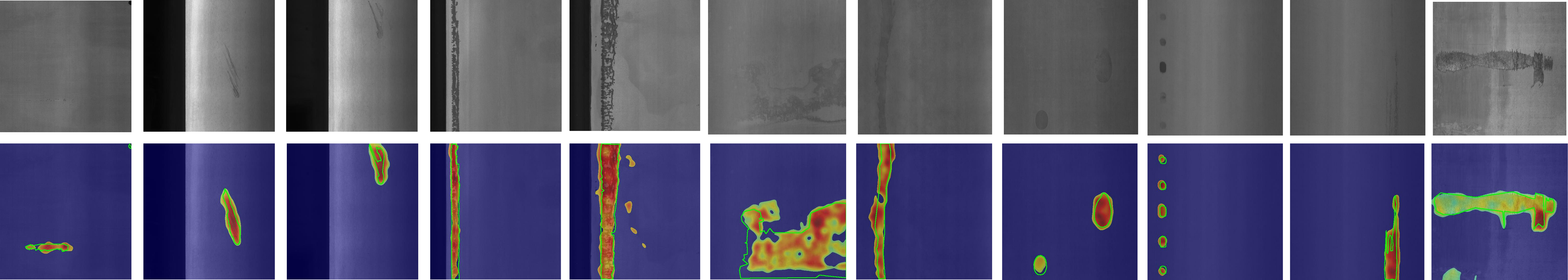}
	\caption{Segmentation results on the GC datasets. The first row depicts the original image, while the second row shows the anomaly segmentation results, with the regions encircled in green representing the ground truth.}
	\label{fig29}
\end{figure}
\begin{figure}[h]
	\centering
	\includegraphics[width=0.96\columnwidth]{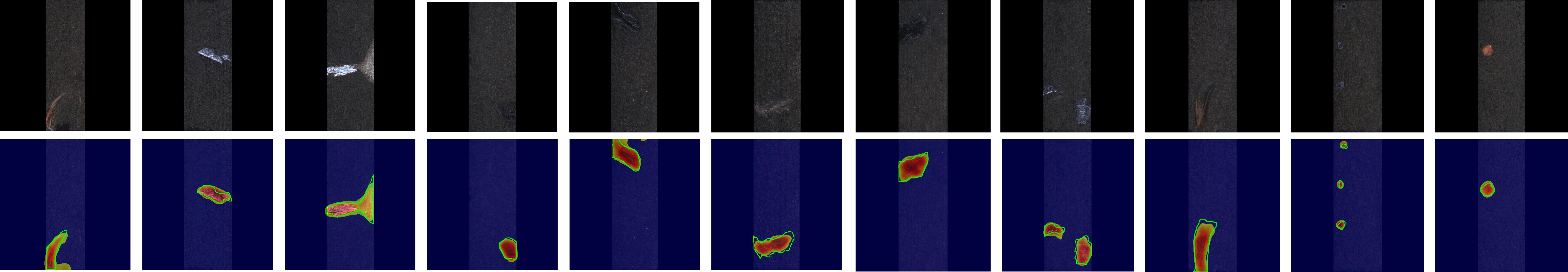}
	\caption{Segmentation results on the KSDD2 datasets. The first row depicts the original image, while the second row shows the anomaly segmentation results, with the regions encircled in green representing the ground truth.}
	\label{fig30}
\end{figure}
\begin{figure}[!h]
	\centering
	\includegraphics[width=0.96\columnwidth]{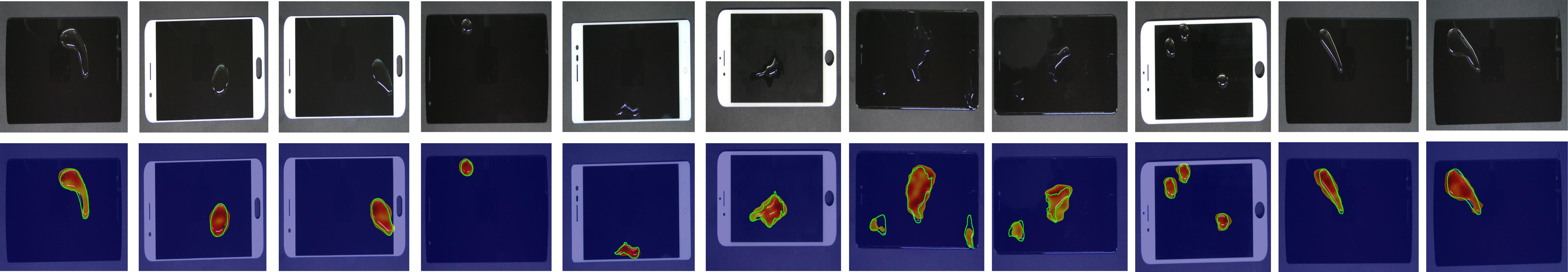}
	\caption{Segmentation results on the MSD datasets. The first row depicts the original image, while the second row shows the anomaly segmentation results, with the regions encircled in green representing the ground truth.}
	\label{fig31}
\end{figure}
\begin{figure}[!h]
	\centering
	\includegraphics[width=0.96\columnwidth]{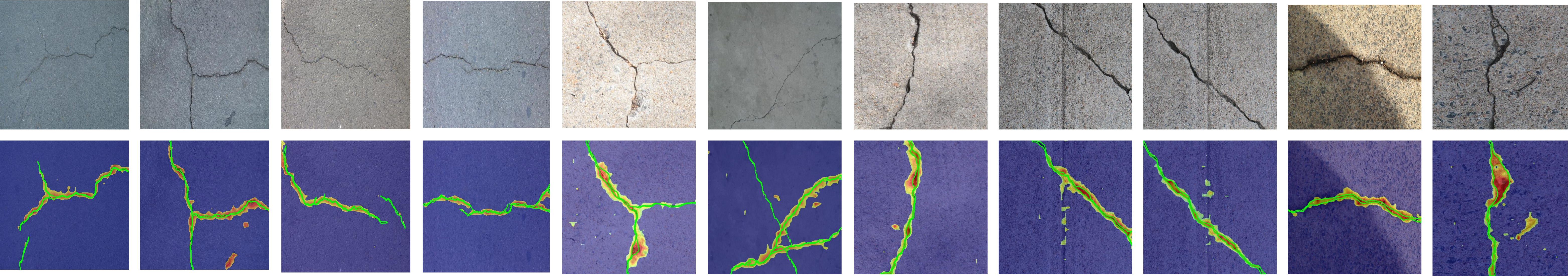}
	\caption{Segmentation results on the Road datasets. The first row depicts the original image, while the second row shows the anomaly segmentation results, with the regions encircled in green representing the ground truth.}
	\label{fig32}
\end{figure}
\begin{figure}[!h]
	\centering
	\includegraphics[width=0.96\columnwidth]{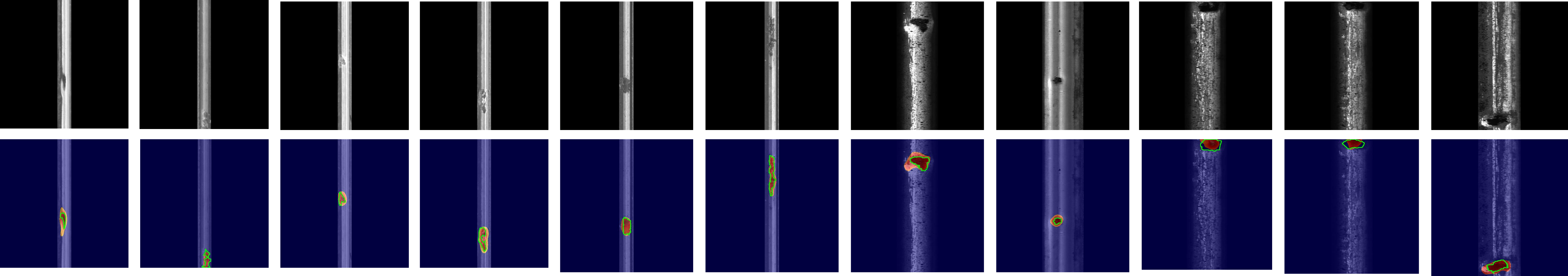}
	\caption{Segmentation results on the RSDD datasets. The first row depicts the original image, while the second row shows the anomaly segmentation results, with the regions encircled in green representing the ground truth.}
	\label{fig33}
\end{figure}
\clearpage
\section{Limitations}   \label{secE}
We have demonstrated the superiority of the proposed VCP-CLIP in the ZSAS task. In this section, we will discuss the two main limitations of our method.
\par 
First, our method can locate the area of the anomaly but may result in some over-detection for minor anomalies, such as the \textit{candle} in Fig. \ref{fig20} and \textit{pipe\_fryum} in Fig. \ref{fig24}. This means that the segmentation results are often slightly larger than the ground truth. This may be attributed to the small input resolution ($336^2$) and large patch size ($14^2$) used in the pretrained backbone (ViT-L-14-336).
\par 
Second, our method cannot accurately localize certain abnormal regions that must rely on normal images for identification, such as pcb4 in Fig. \ref{fig23}. This is because in the ZSAS task setting, VCP-CLIP directly performs anomaly segmentation on novel products without introducing any prior information from normal images. In the future, we plan to further explore the utilization of few-shot techniques to tackle this issue, leveraging the groundwork laid by VCP-CLIP.
\end{document}